\documentclass[11pt]{article}
\pdfoutput=1

\usepackage{microtype}
\usepackage{graphicx}
\usepackage{subfigure}
\usepackage{booktabs} 
\usepackage{amssymb}
\usepackage{bm}
\usepackage{bbm}
\usepackage{amsmath}  
\usepackage{amsthm}
\usepackage{soul}
\usepackage{color, colortbl}
\usepackage{xcolor}
\usepackage{tabularx}
\usepackage{textcomp}
\usepackage{multirow}
\usepackage{mathtools}
\usepackage[margin=1in]{geometry}
\usepackage{titling}
\usepackage{wrapfig}

\usepackage{tikz}
\usetikzlibrary{arrows}
\usetikzlibrary{positioning}

\tikzset{
  treenode/.style = {align=center, inner sep=0pt, text centered,
    font=\sffamily},
  arn_n/.style = {treenode, circle, black, font=\sffamily\bfseries, draw=black,
    fill=white, text width=1.5em},
  arn_r/.style = {treenode, circle, black, font=\sffamily\bfseries, draw=black,
    fill=white, text width=1.0em},
  arn_x/.style = {treenode, rectangle, draw=black,
    minimum width=0.5em, minimum height=0.5em}
}

\usepackage{hyperref}

\usepackage{bm}
\usepackage{enumitem}
\usepackage{caption}

\definecolor{breezy3}{RGB}{139,214,243}
\definecolor{dandelion}{RGB}{255,212,100}

\definecolor{Gray}{gray}{0.9}
\newcolumntype{P}[1]{>{\centering\arraybackslash}p{#1}}

\newcommand*{\ShowNotes}{}

\ifdefined\ShowNotes
  \newcommand{\colornote}[3]{{\color{#1}\bf{#2 #3}\normalfont}}
\else
  \newcommand{\colornote}[3]{}
\fi

\definecolor{darkred}{rgb}{0.7,0.1,0.1}
\definecolor{darkgreen}{rgb}{0.1,0.5,0.1}
\definecolor{cyan}{rgb}{0.7,0.0,0.7}
\definecolor{dblue}{rgb}{0.2,0.2,0.8}
\definecolor{maroon}{rgb}{0.76,.13,.28}
\definecolor{burntorange}{rgb}{0.81,.33,0}
\definecolor{royalpurple}{rgb}{0.47,.31,0.66}

\ifdefined\ShowNotes
  
\else
  
\fi

\newif\ifarxiv

\title{Lost in Transmission: On the Impact of Networking Corruptions on Video Machine Learning Models}

\author{Trenton Chang \\ University of Michigan \\ \texttt{ctrenton@umich.edu} \and Daniel Y. Fu \\ Stanford University \\ \texttt{danfu@cs.stanford.edu}}

\date{}

\begin{document}

\maketitle

\arxivtrue
\nocite{*}

\begin{abstract}
  
We study how \textit{networking corruptions}---data corruptions caused by networking errors---affect video machine learning (ML) models. 
We discover apparent networking corruptions in Kinetics-400, a benchmark video ML dataset. 
In a simulation study, we investigate (1) what artifacts networking corruptions cause, (2) how such artifacts affect ML models, and (3) whether standard robustness methods can mitigate their negative effects.
We find that networking corruptions cause visual and temporal artifacts (i.e., smeared colors or frame drops).
These networking corruptions degrade performance on a variety of video ML tasks, but effects vary by task and dataset, depending on how much temporal context the tasks require.
Lastly, we evaluate data augmentation---a standard defense for data corruptions---but find that it does not recover performance.

\end{abstract}

\begin{figure*}[t]
    \centering
    \includegraphics[width=\linewidth]{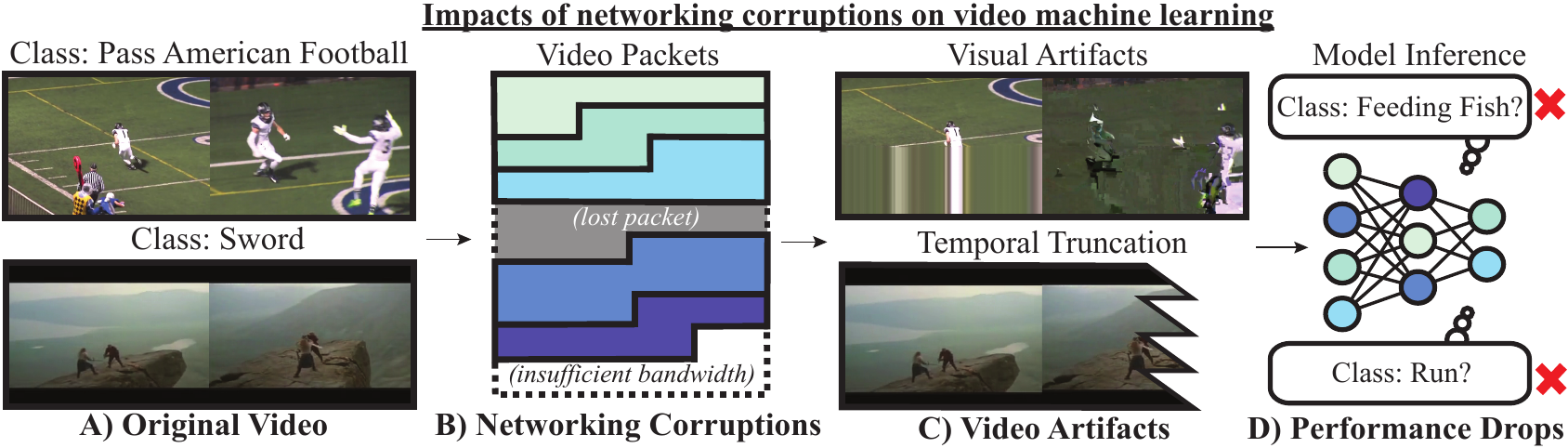}\vspace{-3mm}
    \caption{
    Network issues during video transmission can cause \textit{networking corruptions}, which 
    cause visual artifacts such as color smearing and temporal artifacts such as dropped frames.
    These corruptions degrade the performance of video ML models during inference.
    }\vspace{-5mm}
   \label{fig:main}
\end{figure*}

\section{Introduction}
\label{sec:intro}

Video machine learning (ML) models are known to be vulnerable against data corruptions.
For example, sensor noise (e.g. color shift, blur) and environmental shifts (e.g. snow)~\cite{hendrycks2019benchmarking, hendrycks2019natural} 
can degrade model performance. 
Corruptions due to networking errors during data transmission---which we call \textit{networking corruptions}---are relatively understudied in the literature. 
However, we find previously-undocumented instances of apparent networking corruptions in Kinetics-400, a benchmark video dataset.
These corruptions appear qualitatively different from data corruptions.
Understanding the impacts of such errors may reveal insights on what information is critical for video ML models. This understanding may also be increasingly relevant to video ML deployments, especially those distributed across networks~\cite{zeng2020distream, zeng2021mercury}.

Understanding how networking corruptions impact video ML models is challenging.
To begin, in contrast to data corruptions, networking corruptions can have complex downstream effects on videos.
Networking corruptions can cause artifacts such as color smearing or dropped frames, even when the degree of data corruption is relatively small (Figure~\ref{fig:main}).
Furthermore, video ML models are used for a variety of downstream applications, ranging from action recognition (classifying the action of a single video), to per-frame object segmentation.
As a result, it is unclear how these artifacts impact video ML tasks---for example, smearing artifacts may impact object segmentation differently than action recognition.

In this paper, we study the impact of networking corruptions on video ML models through a simulation study.
First, we qualitatively investigate the impact of networking corruptions on videos (Section~\ref{sec:video}).
We describe two artifact types: (1) visual artifacts, which fill frames with transient visual noise, and (2) temporal truncation, which drops frames from the end of the video. We find that these artifact types are correlated with high packet loss rate and low bandwidth limitations, respectively.

Next, we assess how these artifacts impact four video ML tasks: action recognition, object tracking, object segmentation, and captioning (Section~\ref{sec:models}).
Networking corruptions degrade performance across tasks (up to 46.0\% accuracy on Kinetics-400), but effects vary by task and dataset.
For example, temporal truncation lowers action recognition accuracy, 
but has little impact on object tracking.
We characterize these impacts, and provide guidance on which tasks are affected by which artifacts based on how much temporal context they require.

Lastly, we evaluate whether data augmentation, a standard robustness technique, recovers performance under networking corruptions (Section~\ref{sec:augmentation}). 
Although data augmentation may offer minor protection against strong visual artifacts (+3.0\% accuracy), it has trouble adapting to truncation. We also evaluate a broader suite of defenses (Appendix~\ref{appdx:defenses}), but find that none can fully recover performance.
Targeting specific artifact types is a potential direction for future work.
\subsection{Preliminaries: Video Encoding and Transmission}
\label{sec:prelim}

First, we provide background on video encoding and transmission.

In this paper, we focus on the popular H.264 codec~\cite{richardson2003h264} (we report experiments with an alternate codec in Appendix~\ref{appdx:defenses}). H.264 exploits visual redundancy between frames by storing a few full frames called \textit{keyframes}. The remaining frames store differences with respect to previous frames. 
Frame data is then compressed via quantization and entropy encoding. H.264 also uses error concealment and correction to mitigate bitstream errors~\cite{stockhammer2005h}.

To transmit video, the video bitstream is chunked into network packets that contain frames or partial frames. These are transmitted via a network link, then decoded for downstream use (e.g. ML model inference). Networking corruptions may corrupt/delete portions of the bitstream; even small data losses can cause artifacts or render the video unplayable.

\begin{figure*}[t!]
\centering
\begin{minipage}[t]{.48\textwidth}
\vspace{-0.5mm}
  \centering
  \includegraphics[width=\linewidth]{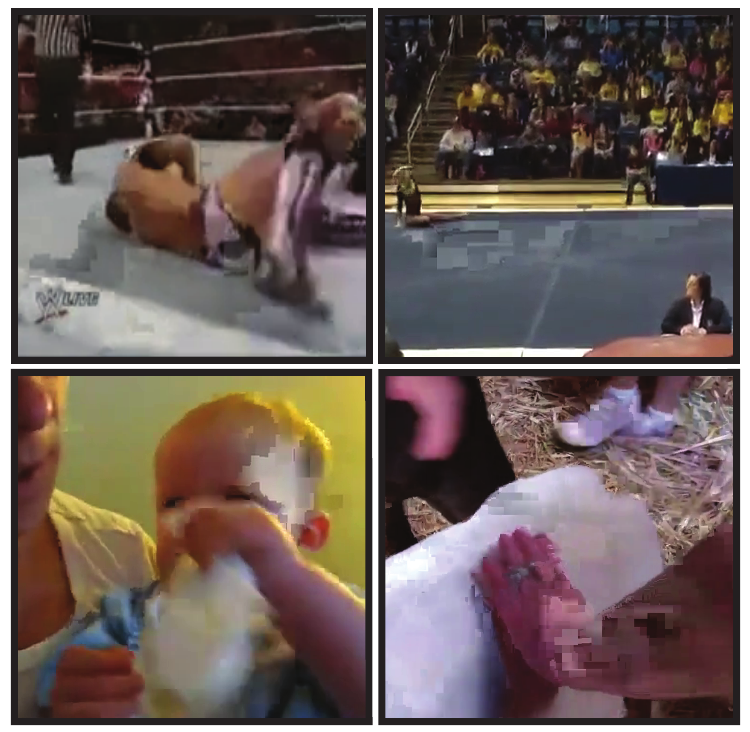}\vspace{-3mm}
  \captionof{figure}{Visual artifacts from apparent networking corruptions in Kinetics-400.}
  \label{fig:natural}
\end{minipage}\hfill%
\begin{minipage}[t]{.48\textwidth}
\vspace{0pt}
  \centering
  \includegraphics[width=\linewidth]{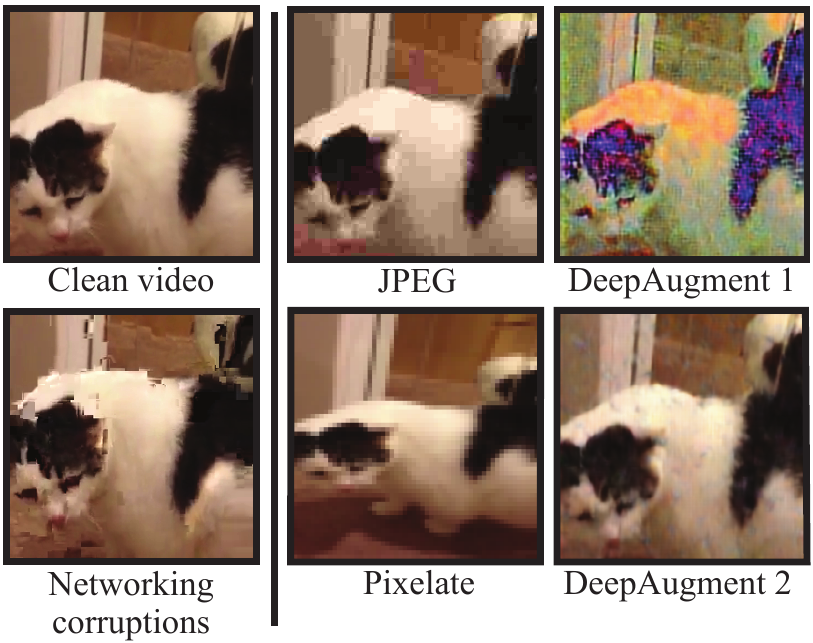}\vspace{-3mm}
  \captionof{figure}{Networking corruptions appear qualitatively different from data corruptions in ImageNet-C and -R \textbf{Top left:} Clean video. \textbf{Bottom left}: Video under networking corruptions (simulated).
  \textbf{Middle:} JPEG compression and pixelation from ImageNet-C.
  \textbf{Right:} Two realizations of DeepAugment~\cite{hendrycks2020faces}.}
  \label{fig:qual_comp}
\end{minipage}
\vspace{-5mm}
\end{figure*}

\section{Networking Corruptions in Kinetics-400}
\label{subsec:corruptions_irl}

We describe apparent networking corruptions found in the Kinetics-400 action recognition dataset.
These corruptions appear as visual artifacts (i.e. blocky pixels, smearing), and differ from corruptions in existing image robustness benchmarks.
Corruptions were found by manually looking for visible artifacts in a random sample of 400 Kinetics-400 videos.

Figure~\ref{fig:natural} shows frames with apparent networking corruptions from Kinetics-400.
Each frame contains ``blocky" artifacts: in a wrestler's shadow (top left), the top of a floor mat (top right), a baby's head (bottom left), and a person's hand (bottom right). 
These artifacts move frame-to-frame, tracking with motion in the video.
This is likely a \emph{keyframe} corruption, which can cause visual artifacts. Applying additional frame differences (i.e., subsequent non-keyframes) would cause the artifacts to move, as observed.

Networking corruptions also appear different from well-studied visual data corruptions. Figure~\ref{fig:qual_comp} compares networking corruptions to those in the ImageNet-C and -R~\cite{hendrycks2020faces, hendrycks2019benchmarking} robustness benchmarks.
While networking corruptions cause local artifacts and visual discontinuities, ImageNet-C and -R corruptions are visually consistent across the frame.
Networking corruption artifacts also change visually over time---a failure mode unique to videos.
\section{Impacts on Videos}
\label{sec:video}

\begin{figure*}[ht!]
    \centering
    \includegraphics[width=\linewidth]{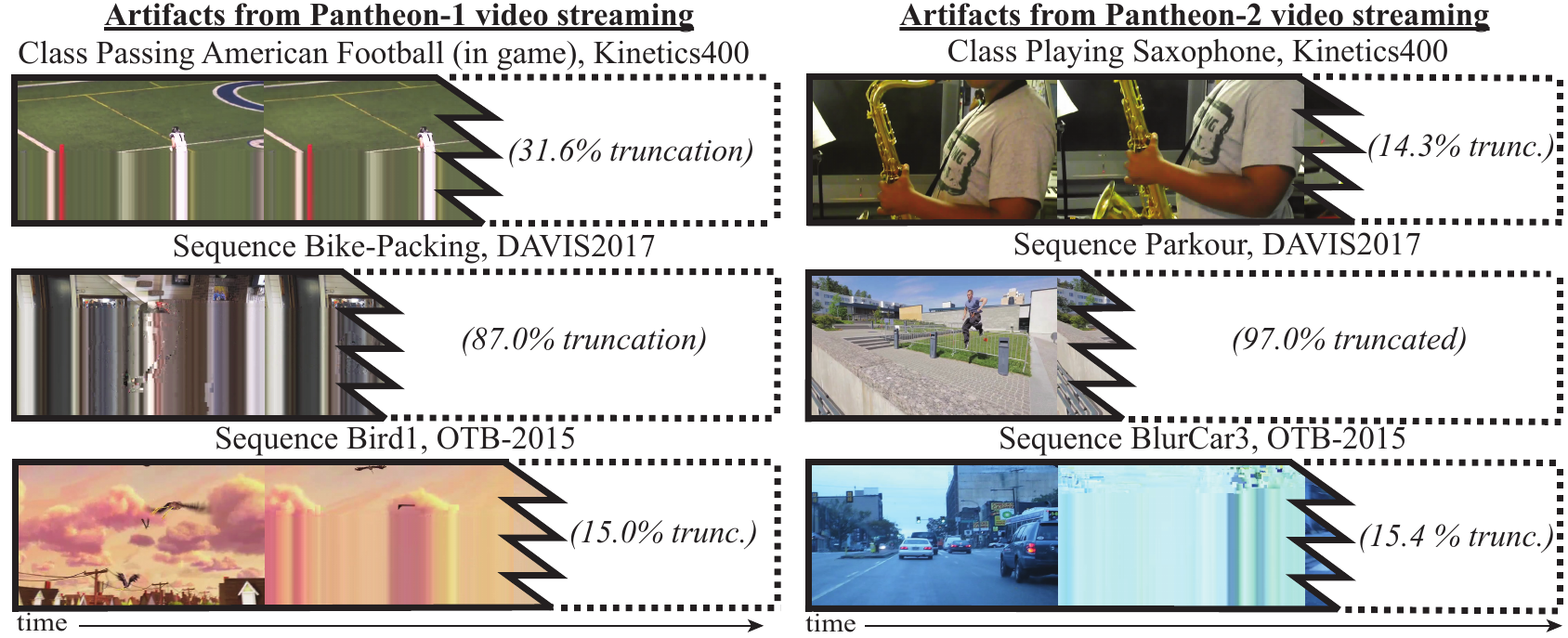}\vspace{-3mm}
    \caption{Visual artifacts and temporal truncation (not to scale) from streaming via Pantheon-1 \textit{(left)} and -2 \textit{(right)} on Kinetics-400 \textit{(top)}, DAVIS \textit{(middle)}, and OTB-2015 \textit{(bottom)}.
    Pantheon-1 causes visual artifacts and temporal truncation. Pantheon-2 typically causes temporal truncation, but occasionally causes visual artifacts (i.e. smears in bottom right).
    }\vspace{-3mm}
    \label{fig:p_examples}
\end{figure*}

\begin{figure*}[ht!]
    \centering
    \includegraphics[width=\linewidth]{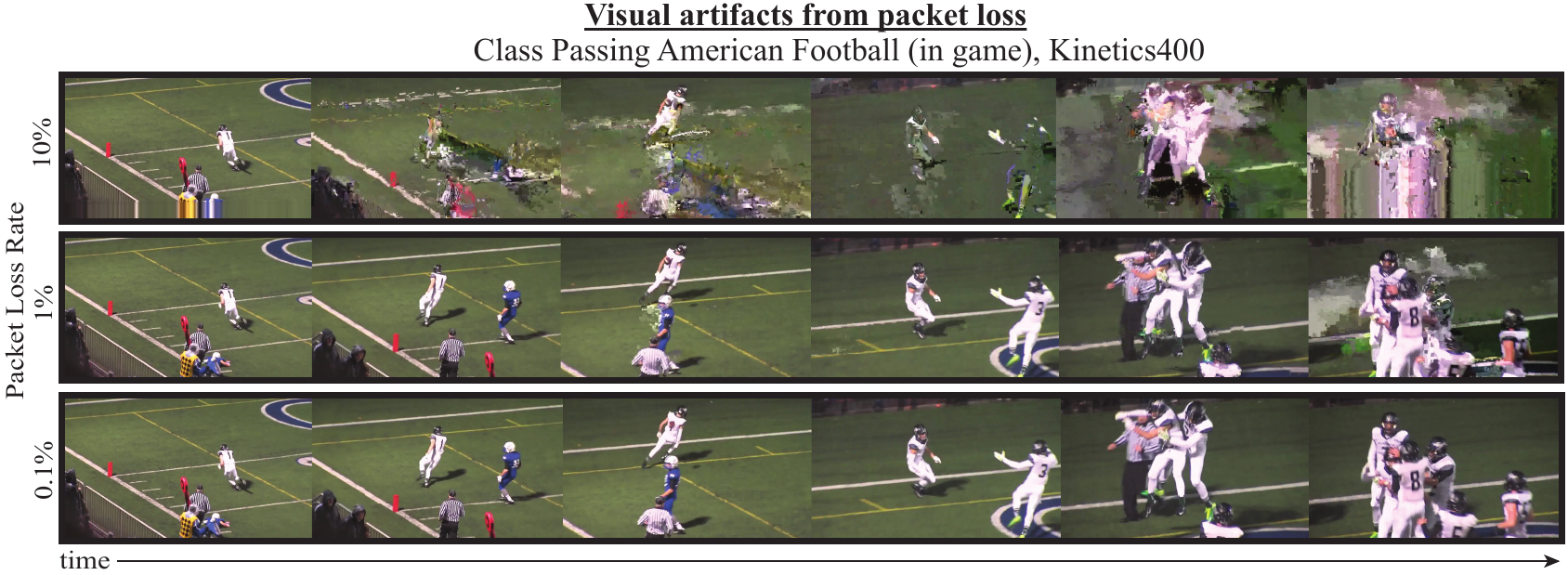}
    \vspace{-3mm}
    \caption{Visual artifacts from a Kinetics-400 video under 10\% \textit{(top)}, 1\% \textit{(middle)}, and 0.1\% \textit{(bottom)} packet loss. Higher packet loss worsens visual artifacts.}\vspace{-5mm}
    \label{fig:packet_loss}
\end{figure*}

Motivated by our empirical findings, we conduct a simulation study of networking corruptions.
We confirm networking corruptions cause visual artifacts similar to those in Kinetics-400.
We also find that networking corruptions cause temporal artifacts (e.g. temporal truncation).
Furthermore, we find that visual and temporal artifacts are correlated with specific network failure modes: high packet loss and low bandwidth.

\textbf{Datasets} We use seven video datasets from four ML tasks with various video lengths, sources (i.e. home video vs. TV), and resolutions. For action recognition, we use HMDB51~\cite{hmdb}, UCF101~\cite{ucf}, Kinetics-400~\cite{kay2017kinetics}, and SomethingSomething-V2 (SSV2)~\cite{goyal2017something}. For single-object tracking, we use OTB-2015~\cite{otb2015}. For video object segmentation, we use DAVIS2017~\cite{davis2017}. For video captioning, we use MSR-VTT~\cite{xu2016msr}.

\begin{table*}
    \centering
    \small
    \caption{Network Link Settings from Pantheon for Simulating Networking Corruptions (left to right: $p$=\% packet loss rate, $B$=bandwidth in Mbps, usage of time-varying packet arrivals, $t$=delay in ms, $S$=queue size in \# of packets).}\vspace{-2mm}
    
    \begin{tabular}{l | c  c c c c c c}
\toprule
        \textbf{Network link name} & $p$ & $B$ (Mbps) & Time-Var. & $t$ (ms) &  $S$ \\
 \midrule
            \textbf{Pantheon-1} (Nepal $\to$ AWS India, Wi-Fi) & 4.77 &  0.57   & Y & 28 & 14 \\
            \textbf{Pantheon-2} (Mexico $\to$ AWS California, Cellular) & 0 & 2.64  & Y & 88  & 130\\
             \textbf{Pantheon-3} (AWS Brazil $\to$ Colombia, Cellular)& 0 & 3.04 & Y & 130 & 426 \\
             \textbf{Pantheon-4} (India $\to$ AWS India, Wired) & 0 & 100.42 & N & 27 & 173\\
\bottomrule
    \end{tabular}
   \label{tab:settings}
\end{table*}

\textbf{Simulation Method}
We stream videos in real-time from one machine to itself via an emulated UDP link. 
Table~\ref{tab:settings} describes the network links we simulate.
These parameters are taken from the Pantheon networking benchmark~\cite{yan-pantheon}, which contains network conditions calibrated to empirically-measured network behavior. 

Each link has five network parameters: (1) packet loss rate $p$ (\%), the \% of packets dropped during transmission, (2) bandwidth $B$ (megabits per second/Mbps), (3) whether packet arrival times are time-varying, (4) delay $t$ (milliseconds), and (5) queue size $S$ (\# of packets), the number of packets the receiver can hold for post-processing before dropping packets.

\begin{wrapfigure}{R}{0.5\textwidth}
    \centering
    \includegraphics[width=\linewidth]{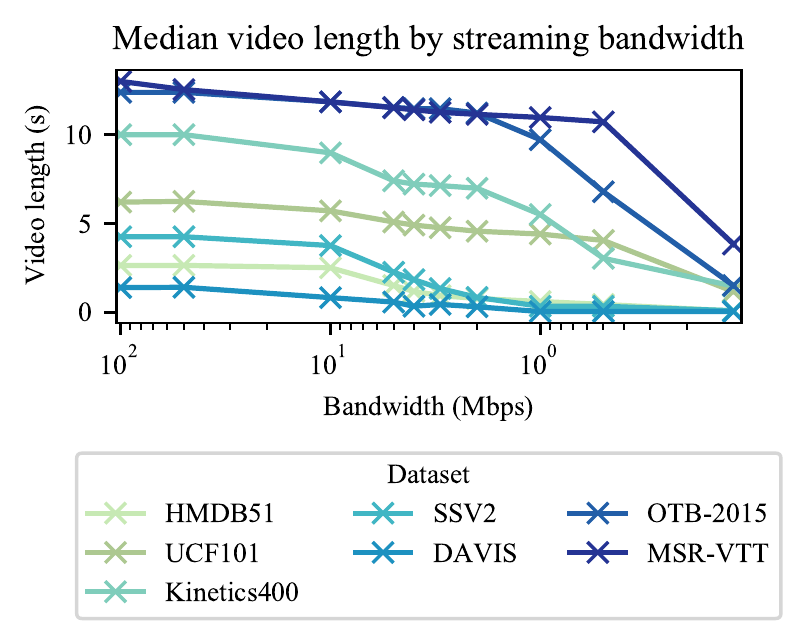}\vspace{-5mm}
    \caption{Median video length \emph{(in seconds)} vs. bandwidth. Bandwidth limits can truncate the video.}\vspace{-5mm}
    \label{fig:length_stats}
\end{wrapfigure}

\subsection{Networking Corruption Artifacts}
\label{subsec:artifacts}

We investigate how networking corruptions impact videos.
We describe two types of artifacts: visual artifacts and temporal truncation. 
Visual artifacts are correlated with high packet loss; temporal truncation is correlated with low bandwidth.

\textbf{Pantheon-1 causes visual artifacts and temporal truncation}
Figure~\ref{fig:p_examples} (left) shows artifacts from Pantheon-1 on HMDB51, DAVIS, and OTB-2015.
Pantheon-1 causes visual artifacts and temporal truncation.
The visual artifacts appear as smearing, color bands, or blocky pixels that change frame-to-frame. 
Such artifacts are likely due to \textit{keyframe} corruptions---errors in encoded full frames. 
Applying encoded frame differences to a corrupted keyframe spreads artifacts to subsequent frames. 
Temporal truncation drops frames from the end of the video.
This disproportionately impacts datasets with shorter videos: median video length drops from 2.67s to 0.60s (-77.5\%) on HMDB51, and from 4.17s to 0.08s (-98.0\%) on SSV2---almost the entire video is lost.
Remaining frames often contain visual artifacts.

\textbf{Pantheon-2 and -3 cause temporal truncation and occasional visual artifacts}
Figure~\ref{fig:p_examples} (right) shows artifacts from Pantheon-2 on HMDB51, DAVIS, and OTB-2015.
In some cases, video frames are missing, but the remaining frames are free of visual artifacts. 
On Kinetics-400, Pantheon-2 and -3 cut median video length from 10.0s to 8.44s (-15.6\%) and 8.51s (-14.9\%) respectively, but the remaining frames appear ``clean" (Figure~\ref{fig:p_examples} top right).
In other cases (particularly OTB-2015), visual artifacts occur in addition to temporal truncation (note color smears, Figure~\ref{fig:p_examples} bottom right).

\textbf{Pantheon-4 sometimes causes minor visual artifacts and temporal truncation}
Pantheon-4 has little impact on videos.
In a sample of 400 videos, small visual artifacts appeared briefly in a minority of videos. 
Temporal truncation is minor; video length drops at most 0.85s (DAVIS).

\textbf{High packet loss is correlated with visual artifacts; low bandwidth is correlated with temporal truncation.}
To investigate why visual artifacts and temporal truncation emerge when streaming, we stream videos via simulated network links in which only packet loss rate and bandwidth vary.
Appendix~\ref{appdx:all_results} reports results for other networking variables; we do not find strong correlations with video artifacts.

High packet loss rate is correlated with visual artifacts (Figure~\ref{fig:packet_loss}),
while limiting bandwidth worsens temporal truncation (Figure~\ref{fig:length_stats}).
This is consistent with the artifacts seen on Pantheon links.
Pantheon-1, which consistently causes visual artifacts, is the only link with a nonzero packet loss rate. Pantheon-1 through -3, which cause temporal truncation, have bandwidths much lower than that of Pantheon-4 (100.42Mbps), where we observed minor truncation.

\begin{table*}[t]
    \centering
    \scriptsize
    \caption{Summary of Impacts of Networking Corruptions}\vspace{-3mm}
    \begin{tabular}{c c | c c}
    \toprule
         \textbf{Task} & \textbf{Output frequency}  & \textbf{Visual artifact impacts} & \textbf{Temporal truncation impacts} \\
    \midrule
        Action Recognition & Per-video & Performance drops & Perf. drops (if temporal context required) \\
        
        Object Tracking & Per-frame & Performance drops & Inflated performance \\
        Object Segmentation & Per-frame & Performance drops & Inflated performance \\
        Video Captioning & Per-video & Small perf. drop; captions change & Captions change \\
    \bottomrule
    \end{tabular}
    \vspace{-3mm}
    \label{tab:findings}
\end{table*}

\begin{figure*}[t]
    \centering
    \includegraphics[width=\linewidth]{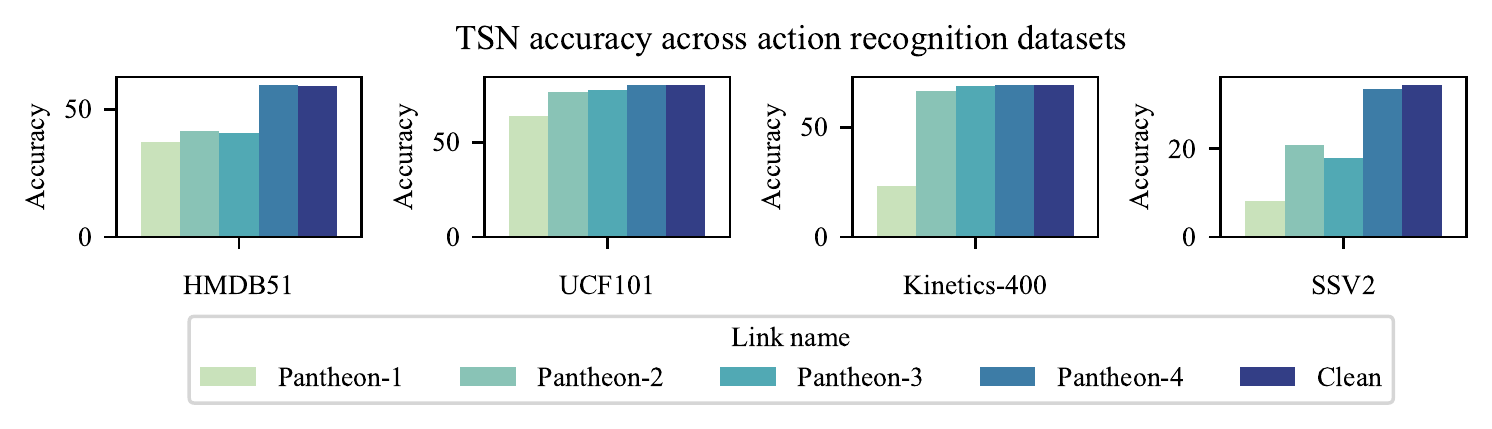}\vspace{-5mm}
    \caption{TSN accuracy on action recognition under networking corruptions and clean data. Networking corruptions hurt model performance across benchmarks.}\vspace{-5mm}
    \label{fig:dataset_diff}
\end{figure*}

\begin{figure*}[t]
    \centering
    \includegraphics[width=\linewidth]{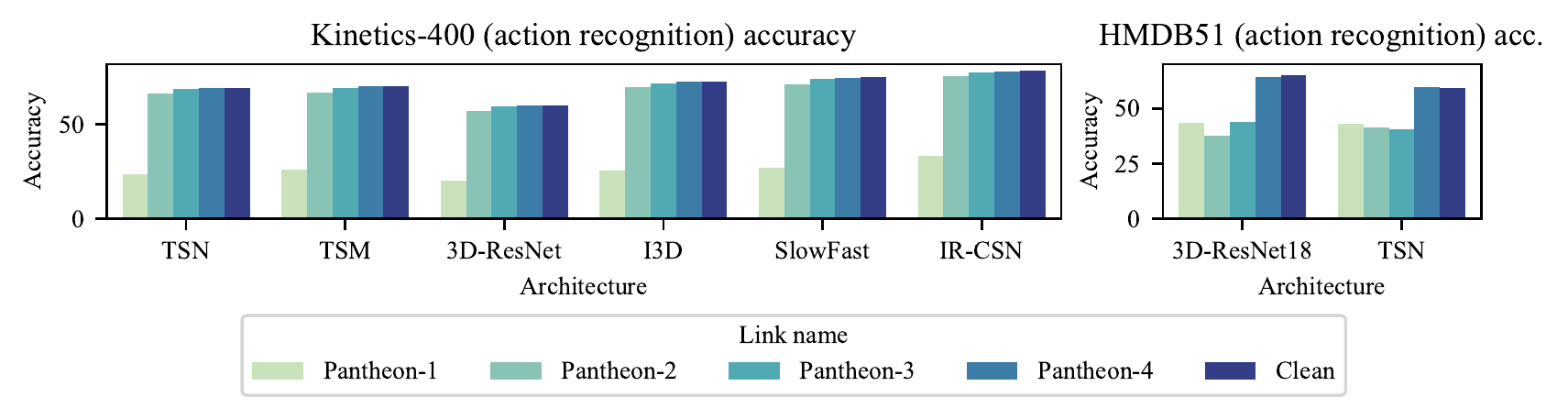}\vspace{-5mm}
    \caption{\textbf{Left:} (from left to right) Accuracy of TSN, TSM, 3D-ResNet, I3D, SlowFast, and IR-CSN on Pantheon-1 to -4 and clean data, Kinetics-400. \textbf{Right:} (from left to right) Accuracy of 3D-ResNet18 and TSN, on Pantheon-1 to -4 and clean data, HMDB51.
    Networking corruptions degrade performance across model architectures.\vspace{-5mm}
    }
    \label{fig:arch}
\end{figure*}

\section{Impacts on Video ML Models}
\label{sec:models}

We assess the impact of networking corruptions on video ML models.
We study models from four tasks: action recognition (Section~\ref{subsec:ar}), object tracking~(Section~\ref{subsec:ot}), object segmentation (Section~\ref{subsec:os}), and video captioning (Section~\ref{subsec:caption}).
These tasks represent various prediction frequencies (one prediction per frame vs. one prediction per video) and modalities (i.e. segmentation mask vs. class prediction).

We summarize our main findings in Table~\ref{tab:findings}. Networking corruptions impact performance on all tasks, but effects vary by task and dataset.
Visual artifacts negatively affect most tasks.
Intuitively, they can obscure objects in a video, destroying action or location information necessary for specific video tasks.
Temporal truncation largely impacts tasks with one prediction per video.
Datasets with shorter videos are more vulnerable: on HMDB51 (action recognition), even minor truncation may remove necessary action information.
Datasets where temporal context is crucial, such as SSV2, are also especially impacted by temporal truncation.
However, segmentation and tracking, which feature frame-by-frame outputs, are less affected, since making the correct prediction on a particular frame requires less temporal context.

\textbf{Evaluation method}
We report standard performance metrics for each task. 
Since networking corruptions can cause videos to become unreadable, 
we report metrics over videos that remain readable after streaming through each link. 
For object tracking \& segmentation, the original annotations may not be aligned with frames from corrupted videos. 
We align ground-truth annotations by matching them to the closest (mean-squared error) frame in the corrupted video. 
Full details are in Appendix~\ref{appdx:all_details}.

\begin{figure*}[t]
    \centering
    \includegraphics[width=\linewidth]{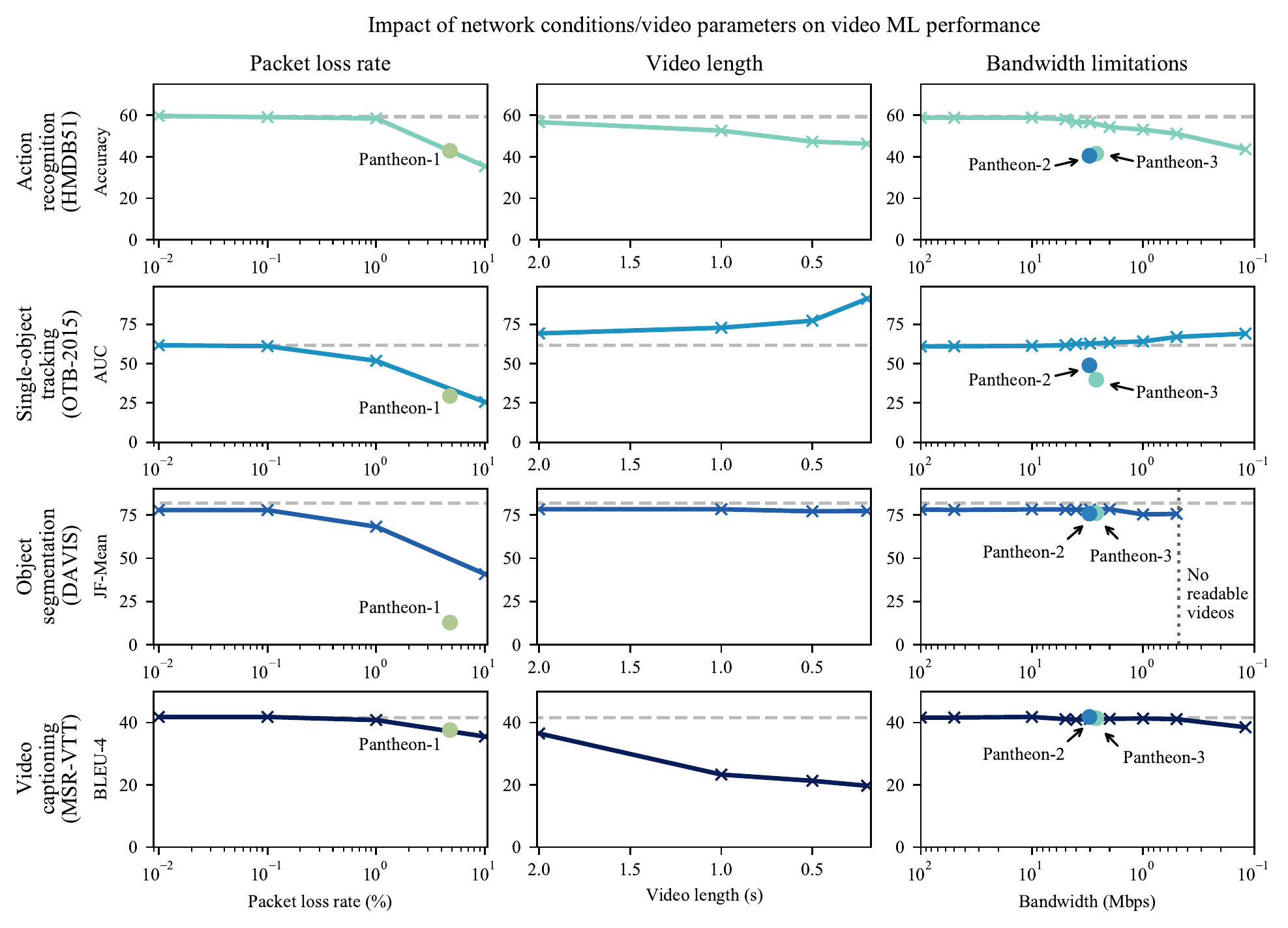}\vspace{-5mm}
    \caption{\textbf{Top:} TSN accuracy (HMDB51, action recognition) under varying packet loss rate \textit{(left)}, video length \textit{(center)}, and bandwidth \textit{(right)}. 
    \textbf{2nd from top:} PrDIMP AUC (OTB-2015, single-object tracking) under varying packet loss rate \textit{(left)}, video length \textit{(center)}, and bandwidth \textit{(right)}.
    \textbf{3rd from top:} DINO JF-Mean (DAVIS, object segmentation) under varying packet loss rate \textit{(left)}, video length \textit{(center)}, and bandwidth \textit{(right)}. 
    \textbf{Bottom:} UniVL BLEU-4 (MSR-VTT, video captioning) under packet loss rate \textit{(left)}, video length \textit{(center)}, and bandwidth \textit{(right)}. 
    }\vspace{-5mm}
    \label{fig:ablations}
\end{figure*}

\begin{figure*}[t]
    \centering
    \includegraphics[width=0.9\linewidth]{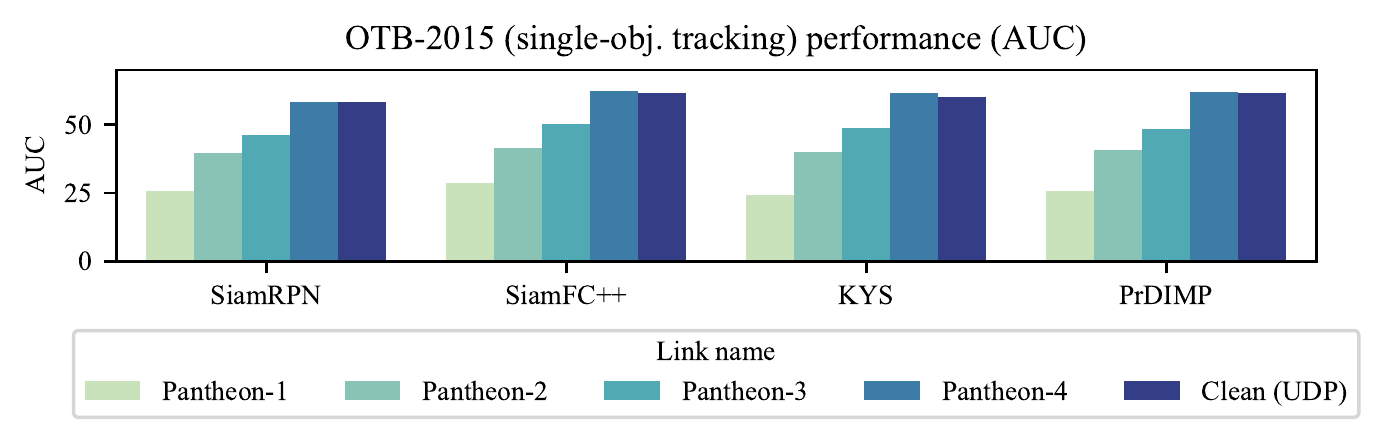}\vspace{-3mm}
    \caption{(from left to right) SiamRPN, SiamFC++, KYS, and PrDIMP performance, OTB-2015, on Pantheon-1 to -4 and a clean link.
    Networking corruptions hurt object tracking performance across multiple architectures and network links.}
    \label{fig:ot}
\end{figure*}

\begin{wrapfigure}{R}{0.5\linewidth}
    \centering
    \includegraphics[width=\linewidth]{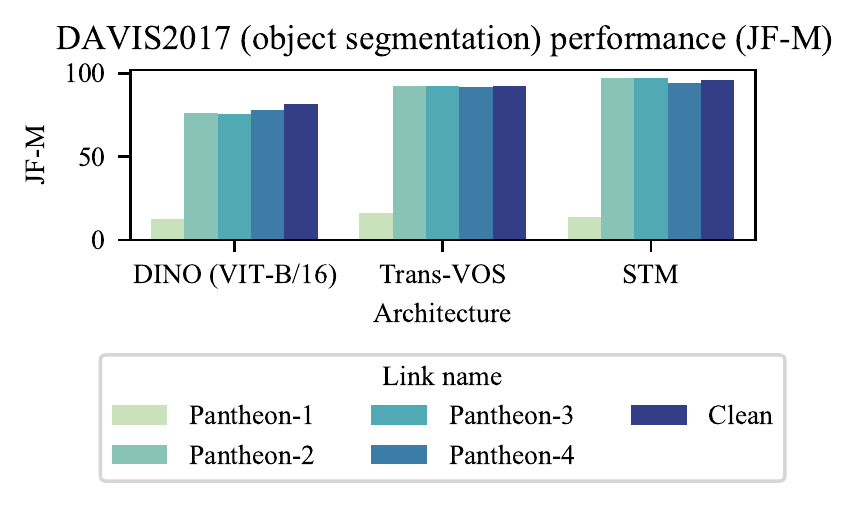}\vspace{-5mm}
    \caption{(from left to right) DINO, Trans-VOS, and STM performance, DAVIS2017, on Pantheon-1 to -4 and clean data. Pantheon-1 degrades performance (strong visual artifacts); other links do not.}
    \label{fig:segmentation}
\end{wrapfigure}

\subsection{Action Recognition}
\label{subsec:ar}

On action recognition, visual artifacts impact all models and datasets, but temporal truncation impacts only certain datasets.

\textbf{Datasets and models}
We evaluate accuracy for a 50-layer TSN~\cite{wang2016temporal} on HMDB51, UCF101, Kinetics-400, and SSV2.
On Kinetics-400, we also evaluate TSM~\cite{lin2019tsm}, 3D-ResNet~\cite{hara2018can}, I3D~\cite{carreira2018quo}, SlowFast~\cite{feichtenhofer2019slowfast} and IR-CSN~\cite{tran2019video}. On HMDB51, 
we also evaluate 3D-ResNet18.

\textbf{Visual artifacts lower performance across action recognition datasets and models}
Figure~\ref{fig:dataset_diff} compares performance on each dataset under networking corruptions. Figure~\ref{fig:arch} shows the performance of various models under networking corruptions.
Over all models and datasets, Pantheon-1 (leftmost bar in each group) lowers accuracy up to 46.0\% (Kinetics-400, 69.4\% to 23.4\%).
Pantheon-2, -3, and -4 do not always lower performance.
This suggests that visual artifacts---common in Pantheon-1---degrade action recognition performance. 

We isolate the effect of visual artifacts by streaming HMDB51 at varying packet loss rates (correlated with visual artifacts). 
Figure~\ref{fig:ablations} (top left) confirms that higher packet loss rate hurts accuracy. 
Thus, visual artifacts from packet loss explain some of the performance drop.\footnote{We examined file checksums/out-of-distribution detection to filter examples with heavy artifacts, but did not use them for our main results as they often discarded clean videos (see Appendix~\ref{appdx:defenses}).}

\textbf{Temporal truncation hurts performance on certain datasets}
Figure~\ref{fig:dataset_diff} shows that Pantheon-2 and -3---which primarily cause temporal truncation---impact HMDB51 and SSV2 accuracy.
These datasets may be vulnerable to truncation.
HMDB51 has short videos (median: 2.67s)---even minor truncation significantly reduces the amount of signal for action recognition.
SSV2 requires more temporal context 
(e.g. object moving to the left), so SSV2 is also more sensitive to truncation.

We isolate the impact of temporal truncation by varying video length and bandwidth limits on HMDB51. 
Figure~\ref{fig:ablations} (top middle) shows that performance drops as video length decreases.
In Figure~\ref{fig:ablations} (top right), performance drops similarly as bandwidth limit decreases.
This pinpoints how much of the performance drop is caused by temporal truncation.\footnote{Empirically, methods to lower bandwidth overhead (i.e. downsampling, switching codec) to mitigate truncation resulted in negative impacts/tradeoffs with ML performance (see Appendix~\ref{appdx:defenses}).}

\subsection{Object Tracking}
\label{subsec:ot}

In summary, object tracking is highly sensitive to visual artifacts, but not truncation.

\textbf{Datasets and models}
We report AUC on OTB-2015 for four object trackers: SiamRPN~\cite{li2018high}, SiamFC++~\cite{xu2020siamfc++}, KYS~\cite{bhat2020know}, and PrDIMP~\cite{danelljan2020probabilistic}.

\textbf{Visual artifacts in Pantheon 1-3 degrade object tracking performance} 
Figure~\ref{fig:ot} shows single-object tracking performance under networking corruptions.
\footnote{Visual artifacts are prevalent on OTB-2015 (even on a clean link). Thus, our baseline is performance on videos streamed via a clean link.}
Networking corruptions on Pantheon-1 to -3 degrade performance, likely due to visual artifacts.
Since objects must be tracked continuously, even small artifacts can disrupt tracking performance by obscuring object(s) of interest. 
As OTB-2015 has longer videos (median: 13.1s\footnote{For datasets with only raw frames (DAVIS \& OTB-2015), we compute video length assuming 30 FPS, a standard frame rate.}), there is more time for visual artifacts to impact tracking.
We confirm this by varying packet loss rate on OTB-2015:
Figure~\ref{fig:ablations} (2nd row, left) shows that tracking AUC drops as packet loss rate exceeds 0.1\%.

\textbf{Object tracking is relatively robust to temporal truncation}
Tracking may be easier on short videos, since visual artifacts have less time to disrupt tracking. 
In fact, temporal truncation can artificially inflate tracking performance.
We confirm this by varying video length on OTB-2015.
Figure~\ref{fig:ablations} (2nd row, middle) shows that performance improves as video length decreases.
As an extreme case, AUC improves by 29.6 when video length drops to 0.2s.
AUC similarly improves as bandwidth decreases (Figure~\ref{fig:ablations}, 2nd row, right).

\subsection{Video Object Segmentation}
\label{subsec:os}

Video object segmentation models are susceptible to visual artifacts, but not temporal truncation---although our evaluation method may be overly lenient.

\begin{table}[t]
\vspace{5mm}
    \centering
    \small
    \caption{MSR-VTT Captioning Performance Under Networking Corruptions} 
    \begin{tabular}{l|ccc}
    \toprule
        \textbf{Link} & \textbf{BLEU-4} & \textbf{ROUGE-L} & \textbf{METEOR}\\
    \midrule
         \textbf{Pantheon-1} & 37.6 & 57.8 &  26.3\\
         \textbf{Pantheon-2} & 41.4 & 60.3 & 28.1\\
         \textbf{Pantheon-3} & \textbf{41.8} & 60.4 & 28.3\\
         \textbf{Pantheon-4} & 41.4 & 60.8 & 28.8 \\
    \midrule
           \rowcolor{Gray} \textbf{Clean} & 41.5 & \textbf{60.9} & \textbf{29.0} \\
    \bottomrule
    \end{tabular}
    \label{tab:captioning}
\end{table}

\begin{table*}[t]
    \centering
    \small
    \caption{Comparison of Generated Captions on MSR-VTT Under Various Networking Corruptions}
    \vspace{-3mm}
    \begin{tabular}{P{0.15\linewidth}|P{0.18\linewidth}P{0.15\linewidth}P{0.15\linewidth}P{0.15\linewidth}}
    \toprule
          \textbf{Clean} & \textbf{Pantheon-1} & \textbf{Pantheon-2} & \textbf{Pantheon-3}   & \textbf{Pantheon-4}  \\
         \toprule
         a man is talking about a small egg  & a man is talking about something & a man is talking about something & a man is talking about something & a man is talking about a product\\
    \midrule
        a minecraft character is walking around & someone is playing a game & someone is playing a game & a minecraft character is walking around & a minecraft character is walking around \\
        
    \midrule
        a man is standing on the stairs and talking & there is a man with cap is walking on the floor & a man is standing & a man is standing & a man is standing on the stairs and talking\\
        
    \bottomrule
    \end{tabular}
    \vspace{-5mm}
    \label{tab:example_captions}
\end{table*}

\textbf{Datasets and Models}
We evaluate DINO~\cite{caron2021emerging}, STM~\cite{oh2019video} and TransVOS~\cite{zhang2020transductive} on DAVIS2017 using JF-mean score.
JF-Mean averages the overlap between the predicted vs. ground-truth mask with mask boundary detection.

\textbf{Visual artifacts impact video object segmentation}
Figure~\ref{fig:segmentation} shows that Pantheon-1 lowers JF-Mean on all methods up to 82.8 points (96.2 to 14.0, STM).
Segmentation may be susceptible to visual artifacts: they can destroy object location information necessary for segmentation.  
We investigate this by varying packet loss levels.
Figure~\ref{fig:segmentation} (third row, left) shows that increasing levels of packet loss negatively impacts performance, suggesting that visual artifacts contribute to the performance drop.

\textbf{Truncation has little impact on object segmentation}
No major performance drops occur on other Pantheon links, or under varying video length and bandwidth (Figure~\ref{fig:segmentation}, third row, middle and right).
Our evaluation may be overly lenient: one clip remains readable on all links (``parkour"), so metrics are reported for that video.
Also, Pantheon-2 and -3 truncate the ``parkour" clip to <0.2s. Thus, models only need to output high-quality masks for a few frames.

\subsection{Video Captioning}
\label{subsec:caption}

On video captioning models, networking corruptions slightly lower standard captioning metrics, but generated captions may change significantly.

\textbf{Datasets and models}
We evaluate UniVL~\cite{luo2020univl} on the MSR-VTT benchmark.
We report three standard metrics for caption quality: BLEU-4, ROUGE-L, and METEOR.
BLEU-4 is the proportion of 4-grams in the generated caption that appear in the reference caption. 
ROUGE-L rewards long common subsequences between generated and reference captions.
METEOR is an F-measure of unigram precision and recall between generated and reference captions, with a penalty for out-of-order words.
Each generated caption is compared to 20 reference captions, yielding 20 scores.
We retain the highest score.

\textbf{Visual artifacts slightly impact video captioning metrics}
Table~\ref{tab:captioning} shows that standard captioning metrics drop slightly on Pantheon-1 (-3.9 BLEU-4, -3.2 ROUGE-L, -2.7 METEOR). Performance slightly drops for the other links.
Intuitively, visual artifacts may increase captioning difficulty by obscuring objects or actions.
We confirm this by varying the amount of packet loss on MSR-VTT.
Figure~\ref{fig:ablations} (bottom left) shows that BLEU-4 drops slightly as packet loss rate exceeds 1\%.

\textbf{Severe truncation lowers video captioning metrics}
Figure~\ref{fig:ablations} (bottom middle) shows that temporal truncation hurts performance at video lengths under 2.0s.
However, MSR-VTT videos are relatively long (Figure~\ref{fig:length_stats}), so performance is stable across bandwidth ranges common in Pantheon (Figure~\ref{fig:ablations}, bottom right).

\textbf{Captioning outputs worsen under video streaming}
Despite small drops in captioning metrics, captions may worsen under video streaming.
Table~\ref{tab:example_captions} compares captions generated under networking corruptions to those generated on clean videos.
We find two failure modes: (1) word or phrase-level substitutions (Table~\ref{tab:example_captions}, 1st+3rd rows) and rephrases (Table~\ref{tab:example_captions}, 2nd row). 
Although captions remain grammatical, these can change or contradict the clean-data caption. 
Up to 74.4\% of captions change after streaming (Pantheon-1); even without corruptions, 48.6\% of captions change.
Thus, video captioning models may still be sensitive to video streaming.
\section{A Baseline Defense}
\label{sec:augmentation}

We investigate whether data augmentation can improve robustness to networking corruptions.
Data augmentation, which aims to improve ML model performance under distribution shifts, has shown promise as a defense for other corruptions~\cite{Dodge2016UnderstandingHI, hendrycks2020faces}.
We find that data augmentation does not recover performance under networking corruptions.
It slightly mitigates severe visual artifacts (+3.0\% lift), but has little effect on temporal truncation.
We evaluate six additional defenses, such as changes to model training/streaming setup and filtering corrupted videos, in Appendix~\ref{appdx:defenses}. These also do not fully restore performance.
Targeting specific artifacts is a potential future direction for improving robustness to networking corruptions.

\textbf{Setup}
We train a 3D-ResNet18 on the HMDB51 action recognition dataset.
During training, each clean video has a 50\% chance to be replaced with network-corrupted video.

\begin{wrapfigure}{R}{0.5\linewidth}
\centering
    \vspace{-2mm}
    \includegraphics[width=\linewidth]{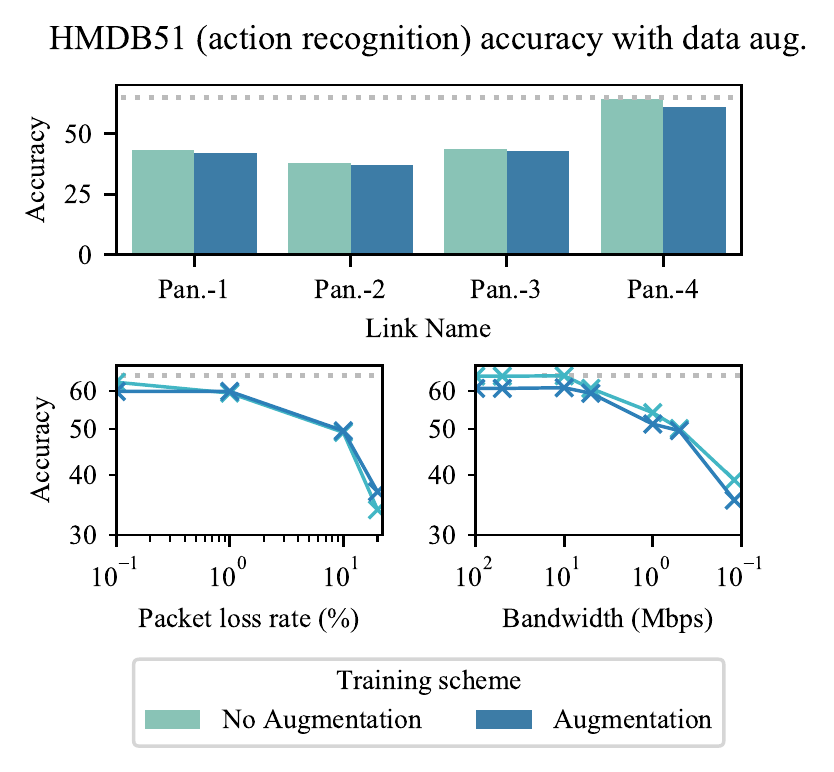}
    \vspace{-5mm}
    \caption{\textbf{Top:} Data augmentation performance using networking corruptions, HMDB51. \textbf{Bottom:} Data augmentation with varying packet loss rates \textit{(left)} or bandwidth limitations \textit{(right)}. Data augmentation does not restore accuracy. It slightly mitigates impacts of high packet loss, but worsens performance at varying bandwidths.}\vspace{-5mm}
    \label{fig:augmentation}
\end{wrapfigure}

\textbf{Data augmentation does not improve performance; shows minor promise for severe visual artifacts}  In Figure~\ref{fig:augmentation} (top), we find that data augmentation does not improve accuracy under networking corruptions. On HMDB51, augmentation improves performance by up to 0.7\% (Pantheon-2), and worsens performance by up to 3.4\% (Pantheon-4). However, Figure~\ref{fig:augmentation} (bottom left) shows that data augmentation may slightly mitigate the impact of severe visual artifacts (packet loss $>10\%$).
At 20\% packet loss, data augmentation restores 3.0\% accuracy, but this lift decreases as packet loss rate decreases.

\textbf{Data augmentation cannot mitigate truncation} 
Figure~\ref{fig:augmentation} (bottom right) shows that data augmentation lowers accuracy under truncation up to 3.7\% (50 \& 100Mbps). 
Intuitively, temporal truncation increases action recognition difficulty, as there is less signal to learn from.
Furthermore, if truncation removes the underlying action from a video, the ground-truth labels become unreliable.
As-is, data augmentation may have trouble adapting to temporal truncation.

\section{Related Work}
\label{sec:related}

\textbf{ML robustness to data corruptions} Recent work establishes benchmarks for varying types of distribution shifts in image data~\cite{geirhos2018imagenettrained, hendrycks2019benchmarking, hendrycks2020faces, koh2021wilds}. 
Our work introduces networking corruptions as an additional failure mode for ML robustness studies. 

\textbf{Networking corruptions \& video} Video corruptions have been studied in networking in the context of video streaming~\cite{frnda2016impact} or codec-level  issues~\cite{boulos2009perceptual, huszak2010analyzing}. 
Network and codec-based solutions to streaming failures are proposed in~\cite{feamster2002packet, kazemi2017joint}. We see our work as complementary, analyzing the impact of networking corruptions in an ML context.

\textbf{Video transmission in ML systems} 
Researchers have proposed live video transmission for ML video analytics due to the cost of video ML model training.
Our work may be of interest to that community.
Distream~\cite{zeng2020distream} uses inter-camera workload-balancing to distribute video processing across multiple nodes.
Spatula~\cite{jain2020spatula} further exploits physical-location correlation of cameras within a cluster.
Chameleon~\cite{jiang2018chameleon} adapts neural network configurations (i.e. frame rate, resolution) to network conditions.

\section{Conclusion}
\label{sec:conclusion}

We study the impact of networking corruptions on video ML.
We discover networking corruptions in Kinetics-400. We then perform a simulation study, describing two artifact types that networking corruptions cause: visual artifacts and temporal truncation.
These corruptions affect video ML models across four video tasks, but effects vary depending on how much temporal context is necessary.
These variable effects reveal what information is important for various video tasks.
Lastly, we find that data augmentation does not mitigate networking corruptions.
We hope our work motivates further studies and defenses for networking corruptions.

\ifarxiv
\section*{Acknowledgments}
We thank Simran Arora, Mayee Chen, Arjun Desai, Karan Goel, Sarah Hooper, Sharon Li, Laurel Orr, Eric Nguyen, Bibek Paudel, Christopher R\'{e}, Nimit Sohoni, Sahaana Suri and Michael Zhang, for their helpful discussions and feedback on early drafts of the paper.
We additionally thank Sadjad Fouladi for the helpful discussions, introducing us to the Mahimahi network simulator/Pantheon benchmark, and providing some computational resources; we also thank Du Tran for assistance with the CSN code repository.
Furthermore, we would also like to acknowledge the FFmpeg community for assisting with patching the RTSP implementation.
We gratefully acknowledge the support of the Department of Defense (DoD) through the National Defense Science and Engineering Graduate Fellowship (NDSEG) Program, the Dean's Office of the Stanford School of Engineering through the Engineering Coterminal Fellowship Program, and members of the Stanford DAWN project: Facebook, Google, and VMWare.

\fi

\bibliographystyle{plain}
\bibliography{main}

\begin{thebibliography}{10}

\bibitem{bhat2020know}
Goutam Bhat, Martin Danelljan, Luc~Van Gool, and Radu Timofte.
\newblock Know your surroundings: Exploiting scene information for object
  tracking.
\newblock In {\em Proceedings of the European Conference on Computer Vision},
  2020.

\bibitem{boulos2009perceptual}
Fadi Boulos, Beno{\^\i}t Parrein, Patrick Le~Callet, and David Hands.
\newblock Perceptual effects of packet loss on {H.264/AVC} encoded videos.
\newblock In {\em Fourth International Workshop on Video Processing and Quality
  Metrics for Consumer Electronics}, 2009.

\bibitem{caba2015activitynet}
Fabian Caba~Heilbron, Victor Escorcia, Bernard Ghanem, and Juan Carlos~Niebles.
\newblock Activity{N}et: A large-scale video benchmark for human activity
  understanding.
\newblock In {\em IEEE/CVF Conference on Computer Vision and Pattern
  Recognition (CVPR)}, 2015.

\bibitem{caron2021emerging}
Mathilde Caron, Hugo Touvron, Ishan Misra, Herv{\'e} J{\'e}gou, Julien Mairal,
  Piotr Bojanowski, and Armand Joulin.
\newblock Emerging properties in self-supervised vision transformers.
\newblock {\em ar{X}iv:2104.14294}, 2021.

\bibitem{carreira2018quo}
Jo{\~{a}}o Carreira and Andrew Zisserman.
\newblock Quo vadis, action recognition? {A} new model and the {K}inetics
  dataset.
\newblock In {\em {IEEE/CVF} Conference on Computer Vision and Pattern
  Recognition (CVPR)}, 2017.

\bibitem{danelljan2020probabilistic}
Martin Danelljan, Luc~Van Gool, and Radu Timofte.
\newblock Probabilistic regression for visual tracking.
\newblock In {\em {IEEE/CVF} Conference on Computer Vision and Pattern
  Recognition (CVPR)}, 2020.

\bibitem{deng2009ImageNet}
Jia Deng, Wei Dong, Richard Socher, Li-Jia Li, Kai Li, and Li~Fei-Fei.
\newblock Imagenet: A large-scale hierarchical image database.
\newblock In {\em IEEE Conference on Computer Vision and Pattern Recognition
  (CVPR)}, 2009.

\bibitem{Dodge2016UnderstandingHI}
Samuel~F. Dodge and Lina Karam.
\newblock Understanding how image quality affects deep neural networks.
\newblock {\em Eighth International Conference on Quality of Multimedia
  Experience (QoMEX)}, pages 1--6, 2016.

\bibitem{feamster2002packet}
Nick Feamster and Hari Balakrishnan.
\newblock Packet loss recovery for streaming video.
\newblock In {\em 12th International Packet Video Workshop}, 2002.

\bibitem{feichtenhofer2019slowfast}
Christoph Feichtenhofer, Haoqi Fan, Jitendra Malik, and Kaiming He.
\newblock Slow{F}ast networks for video recognition.
\newblock In {\em {IEEE} International Conference on Computer Vision (ICCV)},
  2019.

\bibitem{frnda2016impact}
Jaroslav Frnda, Miroslav Voznak, and Lukas Sevcik.
\newblock Impact of packet loss and delay variation on the quality of real-time
  video streaming.
\newblock {\em Telecommunication Systems}, 62(2), 2016.

\bibitem{geirhos2018imagenettrained}
Robert Geirhos, Patricia Rubisch, Claudio Michaelis, Matthias Bethge, Felix~A.
  Wichmann, and Wieland Brendel.
\newblock Image{N}et-trained {CNN}s are biased towards texture; increasing
  shape bias improves accuracy and robustness.
\newblock In {\em 7th International Conference on Learning Representations
  (ICLR)}, 2019.

\bibitem{goyal2017something}
Raghav Goyal, Samira Ebrahimi~Kahou, Vincent Michalski, Joanna Materzynska,
  Susanne Westphal, Heuna Kim, Valentin Haenel, Ingo Fruend, Peter Yianilos,
  Moritz Mueller-Freitag, et~al.
\newblock The ``something something" video database for learning and evaluating
  visual common sense.
\newblock In {\em IEEE International Conference on Computer Vision (ICCV)},
  2017.

\bibitem{hara2018can}
Kensho Hara, Hirokatsu Kataoka, and Yutaka Satoh.
\newblock Can spatiotemporal {3D CNNs} retrace the history of {2D CNNs} and
  {ImageNet}?
\newblock In {\em IEEE/CVF Conference on Computer Vision and Pattern
  Recognition (CVPR)}, 2018.

\bibitem{hendrycks2020faces}
Dan Hendrycks, Steven Basart, Norman Mu, Saurav Kadavath, Frank Wang, Evan
  Dorundo, Rahul Desai, Tyler Zhu, Samyak Parajuli, Mike Guo, Dawn Song, Jacob
  Steinhardt, and Justin Gilmer.
\newblock The many faces of robustness: {A} critical analysis of
  out-of-distribution generalization.
\newblock In {\em IEEE International Conference on Computer Vision (ICCV)},
  2021.

\bibitem{hendrycks2019benchmarking}
Dan Hendrycks and Thomas~G. Dietterich.
\newblock Benchmarking neural network robustness to common corruptions and
  perturbations.
\newblock In {\em International Conference on Learning Representations (ICLR)},
  2019.

\bibitem{hendrycks2019natural}
Dan Hendrycks, Kevin Zhao, Steven Basart, Jacob Steinhardt, and Dawn Song.
\newblock Natural adversarial examples.
\newblock {\em ar{X}iv:1907.07174}, 2019.

\bibitem{huszak2010analyzing}
Arpad Huszák and Sandor {Imre}.
\newblock Analysing {GOP} structure and packet loss effects on error
  propagation in {MPEG-4} video streams.
\newblock In {\em 4th International Symposium on Communications, Control and
  Signal Processing (ISCCSP)}, 2010.

\bibitem{jain2020spatula}
Samvit Jain, Xun Zhang, Yuhao Zhou, Ganesh Ananthanarayanan, Junchen Jiang,
  Yuanchao Shu, Victor Bahl, and Joseph Gonzalez.
\newblock Spatula: Efficient cross-camera video analytics on large camera
  networks.
\newblock In {\em ACM/IEEE Symposium on Edge Computing (SEC)}, 2020.

\bibitem{jiang2018chameleon}
Junchen Jiang, Ganesh Ananthanarayanan, Peter Bodik, Siddhartha Sen, and Ion
  Stoica.
\newblock Chameleon: Scalable adaptation of video analytics.
\newblock In {\em Proceedings of the 2018 Conference of the ACM Special
  Interest Group on Data Communication}, 2018.

\bibitem{kay2017kinetics}
Will Kay, Jo{\~{a}}o Carreira, Karen Simonyan, Brian Zhang, Chloe Hillier,
  Sudheendra Vijayanarasimhan, Fabio Viola, Tim Green, Trevor Back, Paul
  Natsev, Mustafa Suleyman, and Andrew Zisserman.
\newblock The {K}inetics human action video dataset.
\newblock {\em ar{X}iv:1705.06950}, 2017.

\bibitem{kazemi2017joint}
Mohammad Kazemi, Razib Iqbal, and Shervin Shirmohammadi.
\newblock Joint intra and multiple description coding for packet loss resilient
  video transmission.
\newblock {\em IEEE Transactions on Multimedia}, 20(4), 2017.

\bibitem{koh2021wilds}
Pang~Wei Koh, Shiori Sagawa, Henrik Marklund, Sang~Michael Xie, Marvin Zhang,
  Akshay Balsubramani, Weihua Hu, Michihiro Yasunaga, Richard~Lanas Phillips,
  Irena Gao, Tony Lee, Etienne David, Ian Stavness, Wei Guo, Berton~A.
  Earnshaw, Imran~S. Haque, Sara Beery, Jure Leskovec, Anshul Kundaje, Emma
  Pierson, Sergey Levine, Chelsea Finn, and Percy Liang.
\newblock {WILDS}: A benchmark of in-the-wild distribution shifts.
\newblock In {\em International Conference on Machine Learning (ICML)}, 2021.

\bibitem{hmdb}
Hilde Kuehne, Hueihan Jhuang, Estibaliz Garrote, Tomaso Poggio, and Thomas
  Serre.
\newblock {HMDB}: {A} large video database for human motion recognition.
\newblock In {\em IEEE International Conference on Computer Vision (ICCV)},
  2011.

\bibitem{li2018high}
Bo~Li, Junjie Yan, Wei Wu, Zheng Zhu, and Xiaolin Hu.
\newblock High performance visual tracking with siamese region proposal
  network.
\newblock In {\em IEEE/CVF Conference on Computer Vision and Pattern
  Recognition (CVPR)}, 2018.

\bibitem{liang2018enhancing}
Shiyu Liang, Yixuan Li, and R.~Srikant.
\newblock Enhancing the reliability of out-of-distribution image detection in
  neural networks.
\newblock In {\em International Conference on Learning Representations (ICLR)},
  2018.

\bibitem{lin2019tsm}
Ji~Lin, Chuang Gan, and Song Han.
\newblock Tsm: Temporal shift module for efficient video understanding.
\newblock In {\em IEEE/CVF International Conference on Computer Vision (ICCV)},
  2019.

\bibitem{luo2020univl}
Huaishao Luo, Lei Ji, Botian Shi, Haoyang Huang, Nan Duan, Tianrui Li, Jason
  Li, Taroon Bharti, and Ming Zhou.
\newblock {UniVL}: A unified video and language pre-training model for
  multimodal understanding and generation.
\newblock {\em ar{X}iv:2002.06353}, 2020.

\bibitem{netravali2015mahimahi}
Ravi Netravali, Anirudh Sivaraman, Somak Das, Ameesh Goyal, Keith Winstein,
  James Mickens, and Hari Balakrishnan.
\newblock Mahimahi: Accurate record-and-replay for {HTTP}.
\newblock In {\em USENIX Annual Technical Conference (USENIX ATC)}, 2015.

\bibitem{oh2019video}
Seoung~Wug Oh, Joon-Young Lee, Ning Xu, and Seon~Joo Kim.
\newblock Video object segmentation using space-time memory networks.
\newblock {\em IEEE Transactions on Pattern Analysis and Machine Intelligence},
  2019.

\bibitem{davis2017}
Jordi Pont-Tuset, Federico Perazzi, Sergi Caelles, Pablo Arbeláez, Alex
  Sorkine-Hornung, and Luc~Van Gool.
\newblock The 2017 {DAVIS} challenge on video object segmentation.
\newblock {\em ar{X}iv:1704.00675}, 2017.

\bibitem{udp}
John Postel.
\newblock User datagram protocol.
\newblock RFC 768, RFC Editor, 1980.

\bibitem{richardson2003h264}
Iain E.~G. Richardson.
\newblock {\em H.264 and MPEG-4 Video Compression}.
\newblock Wiley, 2003.

\bibitem{RFC7826}
H.~Schulzrinne, A.~Rao, R.~Lanphier, M.~Westerlund, and M.~Stiemerling.
\newblock Real-time streaming protocol version 2.0.
\newblock RFC 7826, RFC Editor, 2016.

\bibitem{simonyan2014very}
Karen Simonyan and Andrew Zisserman.
\newblock Very deep convolutional networks for large-scale image recognition.
\newblock In {\em International Conference on Learning Representations (ICLR)},
  2015.

\bibitem{ucf}
Khurram Soomro, Amir~Roshan Zamir, and Mubarak Shah.
\newblock {UCF101:} {A} dataset of 101 human actions classes from videos in the
  wild.
\newblock {\em ar{X}iv:1212.0402}, 2012.

\bibitem{stockhammer2005h}
Thomas Stockhammer and Miska~M. Hannuksela.
\newblock {H.264/AVC} video for wireless transmission.
\newblock {\em IEEE Wireless Communications}, 12(4), 2005.

\bibitem{tran2019video}
Du~Tran, Heng Wang, Matt Feiszli, and Lorenzo Torresani.
\newblock Video classification with channel-separated convolutional networks.
\newblock In {\em {IEEE} International Conference on Computer Vision (ICCV)},
  2019.

\bibitem{wang2016temporal}
Limin Wang, Yuanjun Xiong, Zhe Wang, Yu~Qiao, Dahua Lin, Xiaoou Tang, and Luc
  Van~Gool.
\newblock Temporal segment networks: Towards good practices for deep action
  recognition.
\newblock In {\em European Conference on Computer Vision (ECCV)}. Springer,
  2016.

\bibitem{wong2020fast}
Eric Wong, Leslie Rice, and J.~Zico Kolter.
\newblock Fast is better than free: Revisiting adversarial training.
\newblock In {\em International Conference on Learning Representations (ICLR)},
  2020.

\bibitem{otb2015}
Yi~Wu, Jongwoo Lim, and Ming-Hsuan Yang.
\newblock Object tracking benchmark.
\newblock {\em IEEE Transactions on Pattern Analysis and Machine Intelligence},
  37(9), 2015.

\bibitem{xu2016msr}
Jun Xu, Tao Mei, Ting Yao, and Yong Rui.
\newblock Msr-vtt: A large video description dataset for bridging video and
  language.
\newblock In {\em IEEE/CVF Conference on Computer Vision and Pattern
  Recognition (CVPR)}, 2016.

\bibitem{xu2020siamfc++}
Yinda Xu, Zeyu Wang, Zuoxin Li, Ye~Yuan, and Gang Yu.
\newblock Siam{FC}++: Towards robust and accurate visual tracking with target
  estimation guidelines.
\newblock In {\em Proceedings of the AAAI Conference on Artificial
  Intelligence}, 2020.

\bibitem{yan-pantheon}
Francis~Y. Yan, Jestin Ma, Greg~D. Hill, Deepti Raghavan, Riad~S. Wahby, Philip
  Levis, and Keith Winstein.
\newblock Pantheon: the training ground for {Internet} congestion-control
  research.
\newblock In {\em {USENIX} Annual Technical Conference ({USENIX} {ATC})}, 2018.

\bibitem{mmaction2019}
Dahua~Lin Yue~Zhao, Yuanjun~Xiong.
\newblock {MMAction}.
\newblock \url{https://github.com/open-mmlab/mmaction}, 2019.

\bibitem{zeng2020distream}
Xiao Zeng, Biyi Fang, Haichen Shen, and Mi~Zhang.
\newblock Distream: Scaling live video analytics with workload-adaptive
  distributed edge intelligence.
\newblock In {\em Proceedings of the 18th Conference on Embedded Networked
  Sensor Systems}, 2020.

\bibitem{zeng2021mercury}
Xiao Zeng, Ming Yan, and Mi~Zhang.
\newblock Mercury: A framework for efficient and elastic on-device distributed
  {DNN} training.
\newblock In {\em ACM Conference on Embedded Networked Sensor Systems
  (SenSys)}, 2021.

\bibitem{zhang2020transductive}
Yizhuo Zhang, Zhirong Wu, Houwen Peng, and Stephen Lin.
\newblock A transductive approach for video object segmentation.
\newblock In {\em IEEE/CVF Conference on Computer Vision and Pattern
  Recognition (CVPR)}, 2020.

\end{thebibliography}

\appendix

\appendix
\onecolumn

\title{Lost in Transmission: The Impact of Networking Corruptions on Video Machine Learning Models (Supplementary Material) }
\author{}
\maketitle

We describe experimental setups in more detail and provide additional experimental results.
In Appendix~\ref{appdx:exp_details}, we describe the complete experimental setup for all experiments, including preprocessing and training details.
In Appendix~\ref{appdx:all_results}, we provide complete results for all our experiments.
In Appendix~\ref{appdx:defenses}, we discuss a selection of additional defenses against networking corruptions, but we do not find any to be fully successful.

\section{Experimental Details}
\label{appdx:exp_details}

\begin{table}[h]
    \centering
    \small
    \caption{Pantheon Network Link Parameters Used for Simulation Study} 
    \begin{tabular}{l | c  c c c c c c}
\toprule
        \textbf{Network link name} & $p$ & $B$ (Mbps) & Time-Var. & $t$ (ms) &  $S$ \\
 \midrule
           \textbf{Pantheon-1} (Nepal $\to$ AWS India, Wi-Fi) & 4.77 &  0.57   & Y & 28 & 14 \\
            \textbf{Pantheon-2} (Mexico $\to$ AWS California, Cellular) & 0 & 2.64  & Y & 88  & 130\\
            \textbf{Pantheon-3} (AWS Brazil $\to$ Colombia, Cellular)& 0 & 3.04 & Y & 130 & 426 \\
             \textbf{Pantheon-4} (India $\to$ AWS India, Wired) & 0 & 100.42 & N & 27 & 173\\
            \textbf{Pantheon-5} (AWS Korea $\to$ China, Wired) & 0.06 & 77.72 & N & 51 & 94 \\
            \textbf{Pantheon-6} (AWS California $\to$ Mexico, Wired)& 0 & 114.68 & N & 45 & 450\\
\bottomrule
    \end{tabular}
    \vspace{-0.5cm}
   \label{tab:allsettings}
\end{table}

\label{appdx:all_details}

\subsection{Video Streaming Details}
\label{appdx:all_pantheon}

\textbf{Streaming.} Videos are streamed in real-time at their native frame rate using the \texttt{-re} flag, and encoded as an H.264 bitstream before transmission. For frame-based datasets, images are converted into a H.264-encoded MP4 video at 25 FPS prior to streaming, then converted back into JPEG images using the highest possible quality settings (\texttt{-qmin} and \texttt{-qmax} set to 1). Some videos become unreadable after streaming. 

\textbf{Link Parameters.} We provide the precise settings for all 6 real-world Pantheon links in Table~\ref{tab:allsettings}.\footnote{As reported at \url{https://pantheon.stanford.edu/measurements/emu/.} } Empirically, we found that testing on Pantheon-1 through -4 was sufficient to demonstrate the variety of impacts that networking corruptions have on video ML models. Specifically, Pantheon-4, -5, and -6 usually had similar impacts on ML model performance. Although we explore the effects of the first four Pantheon links in the main paper, we provide results of our experiments over all six links here for completeness.

\subsection{Evaluation Setup}

\textbf{Preprocessing.}
We use RGB features only. Videos are channel-normalized under the same settings as training, namely, ActivityNet~\cite{caba2015activitynet} normalization for 3D-ResNets and ImageNet~\cite{deng2009ImageNet} normalization for the remaining. Inputs are also scaled to the same spatial size used for training. Likewise, temporal clips are sampled under the same length and sampling rate used in training. For 3D-ResNets, this constitutes $112 \times 112$ pixel videos, while for the other architectures, we use short-edge scaling with short side length $224$px, keeping the original aspect ratio. Temporally, we use 16 frame clips for 3D-ResNets, and 32 frame clips for the remaining architectures after temporal downsampling by a factor of 2 (i.e., every other frame). This serves to match the train-test distributions for these video models as closely as possible, thus isolating networking corruptions as the only distributional shift in our experiments.

\textbf{Annotation alignment.} Empirically, after streaming via UDP, video clips may have missing or duplicate frames\footnote{Lost information in non-keyframes destroys motion information, giving the video an illusion of ``freezing" or duplicate frames.} due to lost packets. As UDP does not guarantee the receipt of outgoing packets, any lost packets are recoverable. Thus, frame-level mismatches may cause off-by-N errors if corrupted videos are directly evaluated with the ground-truth annotations. As a concrete example, one might observe that the 3rd frame of a particular corrupted video is visually identical to the 5th frame of the corresponding clean video, the 4th corrupted frame resembles the 6th clean frame, and so on.\footnote{Illustrative example only. Actual alignments may vary.}

To combat this, we perform an annotation alignment step, which matches ground-truth frames and their corresponding annotations to corrupted-video frames. For every frame, we search for the nearest frames in the clean video as measured by pixel-wise mean-squared error (MSE). As a further optimization, as only a few frames are usually dropped, we search only in the temporal neighborhood of the current frame by looking up to 10 frames forward from the last-matched frame in the clean video, and accepting a frame as a ``match" if its difference falls under a certain MSE threshold. We use an MSE threshold of 100 for DAVIS and OTB-2015. When no match is found, we default to the closest MSE match in that window. We continue matching frames until all corrupt frames have a matching clean frame. Then, we assign as the annotation of the corrupt frame the corresponding annotation of the matched clean frame. Finally, we evaluate the corrupted video sequence using the corresponding annotations.

\textbf{Evaluation Strategy.} On action recognition, for fair comparison during inference, on 2D-backbone architectures, we sample eight frames and apply a ten-crop as in~\cite{wang2016temporal}, and on 3D-backbone architectures, we sample 10 clips and apply a three-crop, following the setup of~\cite{mmaction2019, feichtenhofer2019slowfast}, to emulate fully convolutional testing~\cite{simonyan2014very}. Pretrained model weights are sourced from MMAction~\cite{mmaction2019}\footnote{\url{https://github.com/open-mmlab/mmaction2}}, except for two models: (1) 3D-ResNets, which are either sourced from the original paper authors' code repository~\cite{hara2018can}\footnote{\url{https://github.com/kenshohara/3D-ResNets-PyTorch}} or trained using the setup described below, and (2) CSN, which is sourced from the original authors' code repository~\cite{tran2019video}\footnote{\url{https://github.com/facebookresearch/vmz}}.

For DAVIS, we use the official evaluation repository. For OTB-2015, all repositories use equivalent evaluation procedures within each benchmark. For MSR-VTT, since we test one method in each of those tasks, we use the evaluation method of that repository.

Since videos can become readable during transmission, for comparability across all experiments, we report performance metrics over the subset of videos that remained readable on all six Pantheon links, or the \emph{readable intersection}. For experiments that stream videos on other link settings, such as the experiments on varying network parameters or alternative defenses for networking corruptions, we report performance over the same subset of videos. In experiments where a video from the \emph{readable intersection} is missing, which can happen on experiments that vary networking parameters or the streaming setup, the model's prediction on that video is marked as incorrect. 

\subsection{Training Setup}

\textbf{Action Recognition.} We train 3D-ResNet18 on HMDB51 following the setup of~\cite{hara2018can}. For each video clip, we randomly select a contiguous 16-frame segment. Then, all images are first channel-normalized with respect to the ActivityNet~\cite{caba2015activitynet} mean. Following normalization, we apply a multiscale random crop with aspect ratio one at scales $\{1, 1/2^{1/4}, 1/\sqrt{2}, 1/2^{3/4}, 1/2\}$, choosing one of the four corners + center of the video to crop (as in a five-crop). We then resize all videos to have size $112 \times 112$ pixels. We then train for 50 epochs using a starting learning rate of $0.001$ with weight decay $10^{-5}$, optimizing using stochastic gradient descent. When validation loss saturates for 10 epochs, we decrease learning rate by a factor of 10. For more details, consult the method of~\cite{hara2018can}.

\textbf{Video Captioning.} We train the video captioning model following the setup of UniVL~\cite{luo2020univl}, as well as their method for video feature extraction.\footnote{\url{https://github.com/ArrowLuo/VideoFeatureExtractor}}

\subsection{Ablation Setup}

For each ablation experiment, we select a particular value for either packet loss rate, video length, or bandwidth limitations, and stream videos across an otherwise clean link. For Kinetics-400 and SSV2, due to the size of the datasets and computational cost of simulating networking corruptions, we conduct our ablation study on samples of 2000 and 1000 videos, denoted Kinetics-Mini and SSV2-Mini, respectively. 

\textbf{Packet loss rate.} Ablations on packet loss rate allow us to isolate the impact of visual artifacts. We measure performance on videos subjected to packet loss rates of 0.01\%, 0.1\%, 1\%, 10\%, and 20\%.

\textbf{Video length.} Manual control over video length allows us to stratify by the severity of temporal truncation. We examine performance on videos where only the first 0.2s, 0.5s, 1s, and 2s is streamed.

\textbf{Bandwidth limitations.} Adjusting bandwidth limitations on links allows us to isolate the impact of temporal truncation. We measure performance on videos streamed with bandwidth limitations of 0.12, 0.5, 1, 2, 3, 4, 5, 10, 50, and 100 Mbps.

\subsection{Other}

\textbf{Software.} All development is done on Python using PyTorch 1.7 with Torchvision 0.8.1. Network simulation is done using Mahimahi~\cite{netravali2015mahimahi} with the UDP~\cite{udp} protocol. We use a branch of FFmpeg 4.2 for point-to-point RTSP~\cite{RFC7826} based streaming, with an additional patch to the RTSP streaming implementation. 

\textbf{Hardware.} We use a single V100 GPU for model training and evaluation. Simulating network corruptions can be done on CPU only, though for speed, we used 16-core processors (i.e., an AWS \texttt{c5.4xlarge}) since the simulations are highly parallelizable. 

\textbf{Sourcing models.} We source models from the official code release for the paper when possible, except for action recognition tasks, for which we source models from MMAction~\cite{mmaction2019} and the Facebook Video Model Zoo~\cite{tran2019video} (CSN only).

\section{Full Experimental Results}
\label{appdx:all_results}

In this section, we provide full results for our study on the impact of networking corruptions on video ML models. We include an architectural study, and ablations on packet loss, video length, and bandwidth limitations for all tasks: action recognition (Appendix~\ref{appdx:ar}), object tracking (Appendix~\ref{appdx:ot}), object segmentation (Appendix~\ref{appdx:segmentation}), and captioning (Appendix~\ref{appdx:captioning}). For action recognition, we additionally include a brief study on the impact of model capacity, finding that it does not improve robustness to networking corruptions.
Finally, we discuss additional ablations over network parameters (Appendix~\ref{appdx:additional_ablations}) and provide full statistics for how bandwidth limitations affect video length (Appendix~\ref{appdx:video_length}).

\subsection{Full Results on Action Recognition}
\label{appdx:ar}
\begin{table*}
    \centering
    \small
    \caption{Action Recognition Accuracies by Pantheon Link, Kinetics-400}
    \begin{tabular}{l | c  c c c c c c}
    \toprule
        \textbf{Link} & \textbf{TSN} & \textbf{TSM} & \textbf{3D-ResNet} & \textbf{I3D} & \textbf{SlowFast} & \textbf{IR-CSN} \\
        \midrule
            \textbf{Kinetics-Pantheon-1} & 23.4 & 26.3  & 20.3 &  25.7 & 27.3 & 33.6\\
            \textbf{Kinetics-Pantheon-2} & 66.6 & 67.0  &  57.3 & 69.8 & 71.4 & 75.4\\
            \textbf{Kinetics-Pantheon-3} & 68.6 &   69.4  & 59.4 & 71.9 & 74.1 & 77.5\\
           \textbf{Kinetics-Pantheon-4} & 69.4 & 70.2  &  59.9 & 72.8 & 74.6 &78.0\\
            \textbf{Kinetics-Pantheon-5} & 68.6 &  69.5 & 59.1 & 72.1  & 73.9 & 77.6 \\
            \textbf{Kinetics-Pantheon-6 }&  69.3  & 70.0  & 60.0 & 72.7 & 74.2 &  77.9\\
            
        \midrule
            \rowcolor{Gray}\textbf{Clean} & 69.4 &  70.4 &  60.0 & 72.9 & 75.0 & 78.4\\
        \bottomrule
    \end{tabular}

   \label{tab:all_kinetics}
\end{table*}

\begin{table}
    \centering
    \small
    \caption{Action Recognition Accuracies by Pantheon Link, HMDB51}
    \begin{tabular}{l | c  c c c c c c}
    \toprule
         \textbf{Link}  & \textbf{3D-ResNet18} & \textbf{TSN} \\
        \midrule
           \textbf{HMDB-Pantheon-1} & 43.3 & 42.9  \\
           \textbf{HMDB-Pantheon-2}& 37.7 & 41.5 \\
            \textbf{HMDB-Pantheon-3} & 4 &  40.5 \\
            \textbf{HMDB-Pantheon-4} & 64.2 & 59.6 \\
            \textbf{HMDB-Pantheon-5} & 64.5 &  58.6 \\
            \textbf{HMDB-Pantheon-6} &  64.6  & 59.7  \\
            
            \midrule
            \rowcolor{Gray}\textbf{Clean} & 64.8 &  59.3 \\ 
        \bottomrule
    \end{tabular}

   \label{tab:all_hmdb}
\end{table}

\begin{table}
    \centering
    \small
    \caption{Action Recognition Accuracies by Pantheon Link, UCF101} 
    \begin{tabular}{l | c  c c c c c c}
    \toprule
         \textbf{Link}& \textbf{3D-ResNet18} & \textbf{TSN} \\
        \midrule
           \textbf{UCF-Pantheon-1} & 77.9 & 6 \\
           \textbf{UCF-Pantheon-2} & 82.9 & 76.6 \\
            \textbf{UCF-Pantheon-3} & 82.8 &  77.3  \\
            \textbf{UCF-Pantheon-4} & 86.9 & 80.4\\
            \textbf{UCF-Pantheon-5} & 86.9 &  79.9 \\
            \textbf{UCF-Pantheon-6} & 86.9 & 80.4 \\

        \midrule
            \rowcolor{Gray} \textbf{Clean} & 87.0 &  80.3\\
        \bottomrule
    \end{tabular}

   \label{tab:all_ucf}
\end{table}

\begin{table}
    \centering
    \small
    \caption{Action Recognition Accuracies by Pantheon Link, SSV2} 
    \begin{tabular}{l | c  c c c c c c}
    \toprule
         & \textbf{TSN} & \textbf{TSM} \\
        \midrule
            \textbf{SSV2-Pantheon-1} & 9.7 & 6.5 \\
            \textbf{SSV2-Pantheon-2} & 21.0 & 26.5 \\
            \textbf{SSV2-Pantheon-3} & 18.1 &  14.6  \\
            \textbf{SSV2-Pantheon-4} & 33.4 & 57.5 \\
            \textbf{SSV2-Pantheon-5} & 33.3 &  57.4 \\
            \textbf{SSV2-Pantheon-6} & 33.4 & 57.5 \\

        \midrule
            \rowcolor{Gray}\textbf{Clean} & 34.4 &  57.0\\
        \bottomrule
    \end{tabular}

   \label{tab:all_ssv2}
\end{table}

\textbf{Architectural Study.} We provide full results for our architectural study on Kinetics-400 in Table~\ref{tab:all_kinetics}, HMDB51 in Table~\ref{tab:all_hmdb}, UCF101 in Table~\ref{tab:all_ucf}, and SSV2 in Table~\ref{tab:all_ssv2}. For action recognition, we evaluate 2D-backbone architectures like TSN~\cite{wang2016temporal} and TSM~\cite{lin2019tsm}, and 3D-backbone architectures like 3D-ResNets~\cite{hara2018can}, I3D~\cite{carreira2018quo}, SlowFast~\cite{feichtenhofer2019slowfast}, CSN~\cite{tran2019video}, all with 50 layers.

\textbf{Model Capacity.} We provide full results for our study on the effect of model capacity in Table~\ref{tab:all_backbone} as reported in Section~\ref{subsec:ar}, including results on Pantheon-5 and -6. Again, networking corruptions cause similarly-sized across model sizes, with slightly worse effects on larger-capacity models: on Pantheon-1, performance drops by 36.1\% (3D-ResNet18) to 41.1\% (3D-ResNet101).

\begin{table}
    \centering
    \small
    \caption{3D-ResNet Accuracy on Clean and Corrupted Videos by Architecture Depth ($d$), Kinetics-400}
    \vspace{3mm}
    \begin{tabular}{l | c  c c c c c c}
    \toprule
        \textbf{Link} & $d=18$ & $d=34$ & $d=50$ & $d=101$  \\
        \midrule
             \textbf{Kinetics-Pantheon-1} & 17.4 & 20.5 & 20.3 & 20.1 \\
             \textbf{Kinetics-Pantheon-2} & 51.2 &  57.1& 57.3 & 58.2\\
             \textbf{Kinetics-Pantheon-3} & 53.1 & 59.4 & 59.4 & 60.5\\
            \textbf{Kinetics-Pantheon-4} & 53.4 & 59.7 & 59.9 & 61.2 \\
           \textbf{Kinetics-Pantheon-5} & 52.7 &  59.1 & 59.1 & 60.5 \\
           \textbf{Kinetics-Pantheon-6} & 53.6 &  59.2 & 60.0 & 61.3\\

        \midrule
           \rowcolor{Gray} \textbf{Kinetics-Clean} &53.5 & 59.2  &  59.9  & 61.2 \\
          \bottomrule
    \end{tabular}

   \label{tab:all_backbone}
\end{table}

\begin{table}
    \centering
    \small
    \caption{TSN Accuracy by Packet Loss Rate on HMDB51, UCF101, Kinetics-400, and SSV2}

    \begin{tabular}{l|c c c c}
    \toprule
         \textbf{Packet loss} & \textbf{HMDB51} & \textbf{UCF101} & \textbf{Kinetics-Mini} & \textbf{SSV2-Mini} \\
         \textbf{rate (\%)} & &&\\
         \midrule
        \textbf{0.01} & 59.7 & 80.4 & 67.4 & 33.0\\
        \textbf{0.1} & 59.1 & 79.9 & 66.8 & 33.2\\
        \textbf{1} & 58.5 & 77.1 & 60.4 & 31.3\\
        \textbf{10} & 35.4 & 50.7 & 24.5 & 20.0\\
        \textbf{20} & 24.6 & 32.0 & 12.5 & 11.2 \\
    \midrule
    \rowcolor{Gray}\textbf{Clean} & 59.3 & 80.3 & 69.4 & 34.4\\
    \bottomrule
    \end{tabular}
    \label{tab:ar_plr}
\end{table}

\begin{table}
    \centering
    \small
    \caption{TSN Accuracy by Video Length (s) on HMDB51, UCF101, Kinetics-Mini, and SSV2-Mini}
    \begin{tabular}{l|c c c c}
    \toprule
         \textbf{Length (s)} & \textbf{HMDB51} & \textbf{UCF101} & \textbf{Kinetics-Mini} & \textbf{SSV2-Mini} \\
         \midrule
        \textbf{0.2} & 46.3 & 74.3 & 55.0  & 11.8 \\
        \textbf{0.5} & 47.3 & 75.1 & 56.5 & 17.4 \\
        \textbf{1} & 52.6 & 76.1 & 57.1 & 23.6 \\
        \textbf{2} & 56.7 & 78.3 & 58.7 & 33.4 \\
    \midrule
    \rowcolor{Gray}\textbf{Clean} & 59.3 & 80.3 & 67.2 & 35.5\\
    \bottomrule
    \end{tabular}
    \label{tab:ar_trunc}
\end{table}

\begin{table}
    \centering
    \small
    \caption{TSN Accuracy by Bandwidth on HMDB51, UCF101, Kinetics-Mini, and SSV2-Mini}

    \begin{tabular}{l|c c c c}
    \toprule
         \textbf{Bandwidth} & \textbf{HMDB51} & \textbf{UCF101} & \textbf{Kinetics-Mini} & \textbf{SSV2-Mini} \\
         \textbf{(Mbps)} & & &\\
         \midrule
        \textbf{0.12} & 43.6 & 75.2  & 57.6 & 8.5\\
        \textbf{0.5} & 51.1 & 78.1 & 61.6 & 12.8 \\
        \textbf{1} & 53.1 & 78.5 & 63.6 & 13.6 \\
        \textbf{2} & 54.3 & 79.2 & 65.2 & 16.6\\
        \textbf{3} & 56.6 & 79.2 & 64.7 & 21.9\\
        \textbf{4} & 56.7 & 79.3 & 6 & 25.6\\
        \textbf{5} & 58.0 & 79.2 & 63.5 & 27.8\\
        \textbf{10} & 58.9 & 80.0 & 66.9 & 32.7\\
        \textbf{50} & 58.8 & 80.3 & 67.4 & 32.9 \\
        \textbf{100} & 58.8 & 80.3 & 67.2 & 32.9 \\
    \midrule
    \rowcolor{Gray}\textbf{Clean} & 59.3 & 80.3 & 67.2 & 35.5 \\
    \bottomrule
    \end{tabular}
    \label{tab:ar_bw}
\end{table}

\textbf{Impact of packet loss.} We perform our ablation study on the impact of packet loss on action recognition performance on TSN. Table~\ref{tab:ar_plr} shows that packet loss adversely impacts model performance across all datasets. Performance consistently drops as packet loss rate increases. Intuitively, as visual artifacts tend to worsen with increasing packet loss rate, sufficiently high levels of packet loss may destroy visual information necessary for classification.

\textbf{Impact of video length.} We study the impact of video length on action recognition using TSN as well. Table~\ref{tab:ar_trunc} shows that video truncation to fixed lengths hurts model performance across all datasets. Performance drops as truncation worsens. This result makes intuitive sense: more severe truncation means that more action information necessary for classification may be missing. Interestingly, UCF101 maintains high performance even under truncation.

\textbf{Impact of bandwidth limitations.} Again, we perform our ablation study on the impact of bandwidth limitations on action recognition performance on TSN. Table~\ref{tab:ar_bw} shows that bandwidth limitations can hurt model performance across all datasets. Performance worsens as bandwidth limitations decrease further. However, datasets with shorter videos, such as HMDB51 and SSV2-Mini, which suffer up to a 24.4\% accuracy drop (SSV2, 0.12Mbps) are much more severely impacted than datasets with longer videos, such as Kinetics-Mini and UCF101, which only suffer up to a 9.6\% drop (Kinetics-Mini, 0.12Mbps). Intuitively, temporal truncation will disproportionately impact datasets with shorter videos: in shorter videos, truncation is more likely to destroy or cut into action information necessary for identifying an action.

\subsection{Full Results on Video Object Tracking}
\label{appdx:ot}
\textbf{Architectural study.} We present the full results of our architectural study on video objecting in the OTB2015 single-object tracking benchmark in Table~\ref{tab:otb_all}. For single-object tracking, we evaluate SiamRPN~\cite{li2018high}, SiamFC++~\cite{xu2020siamfc++}, KYS~\cite{bhat2020know}, and PrDIMP~\cite{danelljan2020probabilistic}. Because of the unusually high number of visual artifacts attributable to the usage of UDP, we report performance for data streamed on a clean link with UDP (Clean-UDP) as well.

\textbf{Impact of packet loss.} We evaluate the impact of packet loss rate on object tracking performance using PrDIMP for single-object tracking. Table~\ref{tab:ot_plr} shows that increasing packet loss hurts tracking performance. Intuitively, visual artifacts, which emerge in the presence of packet loss, can disrupt tracking performance by destroying information about object location and shape across multiple frames.

\textbf{Impact of video length.} We quantify the effects of video length on object tracking performance, again using PrDIMP. Table~\ref{tab:ot_length} shows that as video length decreases, single-object tracking performance actually increases. As intuition, shorter videos are easier examples for object tracking, because objects in such clips only need to be tracked successfully for fewer frames. The longer the video, the more likely that subtle visual artifacts will disrupt tracking performance. As previously discussed, OTB-2015 is composed of relatively longer videos (median length 13.1s), exacerbating this issue. 

\textbf{Impact of bandwidth limitations.} We study the effects of bandwidth limits on object tracking performance on PrDIMP. Table~\ref{tab:ot_bw} shows that as the bandwidth limit decreases, object tracking performance improves. This is likely since decreasing bandwidth is correlated with shorter videos. This result is consistent with our ablation on video length.

\begin{table}
\centering
    \small
    \caption{Object Tracking Performance (AUC) by Pantheon Link, OTB-2015}
    \begin{tabular}{l | c c c c}
    \toprule
        \textbf{Link} & \textbf{SiamRPN} & \textbf{SiamFC++} & \textbf{KYS} & \textbf{PrDIMP}\\
    \midrule
        \textbf{OTB2015-Pantheon-1} & 25.9  & 28.8  &  24.4 & 25.6 \\
        \textbf{OTB2015-Pantheon-2} & 39.6 & 41.5 & 39.8 & 40.5\\
        \textbf{OTB2015-Pantheon-3} & 46.4 & 50.3 & 48.9 & 48.5 \\
        \textbf{OTB2015-Pantheon-4} & 58.3 & 62.3 & 61.5 & 62.0 \\
         \textbf{OTB2015-Pantheon-5} & 57.6 & 60.2 & 60.0 & 61.0 \\
        \textbf{OTB2015-Pantheon-6} & 57.7 & 61.5 & 60.9 & 61.1 \\
        
    \midrule
        \rowcolor{Gray}\textbf{Clean-UDP} & 58.3 &  61.9 & 60.8 & 61.6\\
        \rowcolor{Gray}\textbf{Clean} & 64.8 & 68.4 & 69.5 & 70.5\\
    \bottomrule
    \end{tabular}

    \label{tab:otb_all}
\end{table}

\begin{table}
    \centering
    \small
    \caption{PrDIMP AUC by Packet Loss Rate, OTB-2015}\vspace{-3mm}
    \begin{tabular}{l|c}
    \toprule
         \textbf{Packet loss} & \textbf{PrDIMP}  \\
         \textbf{rate (\%)} & \\
         \midrule
        \textbf{0.01} & 61.7\\
        \textbf{0.1} & 61.0\\
        \textbf{1} & 51.8\\
        \textbf{10} & 25.6\\
        \textbf{20} & 16.4 \\
    \midrule
    \rowcolor{Gray}\textbf{Clean-UDP} & 61.6\\
    \rowcolor{Gray}\textbf{Clean} & 70.5\\
    \bottomrule
    \end{tabular}
    \label{tab:ot_plr}
\end{table}

\begin{table}
    \centering
    \small
    \caption{PrDIMP AUC by Video Length (s), OTB-2015}\vspace{-3mm}
    \begin{tabular}{l|c}
    \toprule
         \textbf{Length (s)} & \textbf{PrDIMP} \\
         \midrule
        \textbf{0.2} &  91.3 \\
        \textbf{0.5} & 78.5 \\
        \textbf{1} & 75.7\\
        \textbf{2} & 72.4 \\
    \midrule
    \rowcolor{Gray}\textbf{Clean-UDP} & 61.6 \\
    \rowcolor{Gray}\textbf{Clean} & 70.5 \\
    \bottomrule
    \end{tabular}
    \label{tab:ot_length}
\end{table}

\begin{table}
    \centering
    \small
    \caption{PrDIMP AUC by Bandwidth, OTB-2015}

    \begin{tabular}{l|c}
    \toprule
         \textbf{Bandwidth} & \textbf{PrDIMP} \\
         \textbf{(Mbps)} & \\
         \midrule
        \textbf{0.12} & 69.1 \\
        \textbf{0.5} & 67.0\\
        \textbf{1} & 64.2 \\
        \textbf{2} & 63.4 \\
        \textbf{3} & 62.8 \\
        \textbf{4} & 62.6 \\
        \textbf{5} & 61.7 \\
        \textbf{10} & 61.3 \\
        \textbf{50} & 61.0\\
        \textbf{100} & 60.9 \\
    \midrule
    \rowcolor{Gray}\textbf{Clean-UDP} & 61.6 \\
    \rowcolor{Gray}\textbf{Clean} & 70.5 \\
    \bottomrule
    \end{tabular}
    \label{tab:ot_bw}
\end{table}

\begin{table}
    \centering
    \small
    \caption{Segmentation Performance by Pantheon Link, DAVIS} 
    \begin{tabular}{l | c c c}
    \toprule
        \textbf{Link} & \textbf{DINO} & \textbf{STM} & \textbf{TransVOS}\\
    \midrule
        \textbf{DAVIS-Pantheon-1} & 12.8  & 14.0 & 16.1\\
        \textbf{DAVIS-Pantheon-2} & 76.0 & 97.1 & 92.6\\
        \textbf{DAVIS-Pantheon-3} & 75.8 & 97.1 & 92.6\\
        \textbf{DAVIS-Pantheon-4} & 77.8 & 94.1 & 91.7\\
         \textbf{DAVIS-Pantheon-5} &73.1 & 88.4 & 82.2\\
        \textbf{DAVIS-Pantheon-6} & 77.8 & 94.1 & 91.7\\
    \midrule
        \rowcolor{Gray}\textbf{Clean} & 81.7 & 96.3 & 92.3\\
    \bottomrule
    \end{tabular}

    \label{tab:all_davis}
\end{table}

\begin{table}
    \centering
    \small
    \caption{DINO JF-Mean by Packet Loss Rate, DAVIS}

    \begin{tabular}{l|c}
    \toprule
         \textbf{Packet loss} & \textbf{DINO}  \\
         \textbf{rate (\%)} & \\
         \midrule
        \textbf{0.01} & 77.8\\
        \textbf{0.1} & 77.8\\
        \textbf{1} & 68.2\\
        \textbf{10} & 40.8\\
        \textbf{20} & 7.1\\
    \midrule
    \rowcolor{Gray}\textbf{Clean} & 81.7\\
    \bottomrule
    \end{tabular}
    \label{tab:davis_plr}
\end{table}

\begin{table}
    \centering
    \small
    \caption{DINO JF-Mean by Video Length (s), DAVIS} 
    \begin{tabular}{l|c }
    \toprule
         \textbf{Length (s)} & \textbf{DINO} \\
         \midrule
        \textbf{0.2} & 77.3 \\
        \textbf{0.5} & 77.1 \\
        \textbf{1} & 78.3 \\
        \textbf{2} & 78.3 \\
    \midrule
    \rowcolor{Gray}\textbf{Clean} & 81.7 \\
    \bottomrule
    \end{tabular}
    \label{tab:davis_length}
\end{table}

\begin{table}
    \centering
    \small
    \caption{DINO JF-Mean by Bandwidth, DAVIS} 

    \begin{tabular}{l|c}
    \toprule
         \textbf{Bandwidth} & \textbf{DINO} \\
         \textbf{(Mbps)} & \\
         \midrule
        \textbf{0.12} & N/A \\
        \textbf{0.5} & 75.6 \\
        \textbf{1} & 75.3\\
        \textbf{2} & 78.3 \\
        \textbf{3} & 78.2 \\
        \textbf{4} & 78.2 \\
        \textbf{5} & 78.2 \\
        \textbf{10} & 77.2 \\
        \textbf{50} & 77.9\\
        \textbf{100} &  78.1\\
    \midrule
    \rowcolor{Gray}\textbf{Clean} & 81.7 \\
    \bottomrule
    \end{tabular}
    \label{tab:davis_bw}
\end{table}

\subsection{Full Results on Video Object Segmentation}
\label{appdx:segmentation}
\textbf{Architectural study.} We present the full results for our architectural study on the DAVIS2017 video object segmentation benchmark in Table~\ref{tab:all_davis}. For segmentation, we evaluate DINO~\cite{caron2021emerging}, STM~\cite{oh2019video}, and Trans-VOS~\cite{zhang2020transductive}. For DINO in particular, we use the base vision transformer setting (ViT-B/16) with patch size 16 due to memory constraints.

\textbf{Impact of packet loss.} We study the impact of packet loss on object segmentation on DAVIS using the DINO architecture. Table~\ref{tab:davis_plr} shows that increasing packet loss consistently hurts model performance in terms of JF-Mean score. Intuitively, visual artifacts make it more difficult to output a correct segmentation mask, as the artifacts destroy information about object location and boundaries.

\textbf{Impact of video length.} We study the impact of video length on object segmentation on DAVIS using the DINO architecture. Table~\ref{tab:davis_length} shows that constricting video length slightly decreases segmentation performance up to 4.4 JF-Mean points. Intuitively, since segmentation mask predictions depend only on the current and previous frames, temporal truncation is unlikely to affect performance significantly.

\textbf{Impact of bandwidth limitations.} We study the impact of bandwidth limitations on object segmentation on DAVIS using the DINO architecture. We report results with the caveat that at 0.12Mbps, the set of readable videos was completely disjoint from the intersection of readable videos under Pantheon streaming. We denote this situation with the annotation ``N/A." Table~\ref{tab:davis_bw} shows that as the bandwidth limit decreases, performance fluctuates, decreasing up to 6.1 JF-Mean points. This result is consistent with the ablation on video length. Ultimately, temporal truncation is unlikely to impact segmentation performance severely.

\subsection{Full Results on Video Captioning}
\label{appdx:captioning}
\textbf{Architectural study.} We present full results for video captioning on the MSR-VTT benchmark in Table~\ref{tab:msrvtt_all}. We evaluate UniVL under the video-only setup as described in~\cite{luo2020univl}.

\textbf{Impact of packet loss.} We isolate the impact of packet loss on video captioning performance. Interestingly, as shown in Table~\ref{tab:captioning_plr}, packet loss does not severely impact captioning performance until packet loss rates exceed $10\%$. 
At high levels of packet loss, performance in terms of BLEU-4, ROUGE-L, and METEOR drops (-10.5 BLEU-4, -8.1 ROUGE-L, -6.1 METEOR at 20\% packet loss rate). The performance drop makes intuitive sense: sufficient visual artifacts in  corrupted video may destroy information relevant to the caption.

\textbf{Impact of video length.} We evaluate the impact of video length on video captioning performance.
Table~\ref{tab:captioning_length} shows that as video length decreases beyond 2s, captioning performance worsens. In fact, when truncated to 2s, captioning performance is already slightly degraded (-4.9 BLEU-4, -4.5 ROUGE-L, -3.8 METEOR). However, performance worsens as video length decreases: at 0.2s, performance drops by 21.8 BLEU-4, 12.6 ROUGE-L, and 9.2 METEOR points. The negative impact of truncation on video captioning makes intuitive sense: if truncation erases parts of the video with important objects or actions in the scene, the captioning task becomes more difficult, or even impractical.

\textbf{Impact of bandwidth limitations.} We evaluate the impact of bandwidth limits on video captioning performance. Table~\ref{tab:captioning_bw} shows that at bandwidth limits tested, video captioning performance drops trivially, with 0.12Mbps resulting in the worst performance drop (-2.9 BLEU-4, -2.7 ROUGE-L, -2.6 METEOR). This suggests that at the bandwidth limits in our ablation study, the temporal truncation is insufficient to adversely impact performance. 
This is expected as MSR-VTT features relatively long videos (median length 13.04s).

\begin{table}
    \centering
    \small
    \caption{UniVL Captioning Performance by Pantheon Link, MSR-VTT}
    \begin{tabular}{l|ccc}
    \toprule
        \textbf{Link} & \textbf{BLEU-4} & \textbf{ROUGE-L} & \textbf{METEOR}\\
    \midrule
         \textbf{MSR-VTT-Pantheon-1} & 37.6 & 57.8 &  26.3\\
         \textbf{MSR-VTT-Pantheon-2} & 41.4 & 60.3 & 28.1\\
         \textbf{MSR-VTT-Pantheon-3} & 41.8 & 60.4 & 28.3\\
         \textbf{MSR-VTT-Pantheon-4} & 41.4 & 60.8 & 28.8 \\
         \textbf{MSR-VTT-Pantheon-5} & 41.3 & 60.7 & 28.8 \\
         \textbf{MSR-VTT-Pantheon-6} & 41.5 & 60.8 & 28.8 \\
    \midrule
           \rowcolor{Gray} \textbf{Clean} & 41.5 & 60.9 & 29.0 \\
    \bottomrule

    \end{tabular}

    \label{tab:msrvtt_all}
\end{table}

\begin{table}
    \centering
    \small
    \caption{UniVL Captioning Performance by Packet Loss Rate, MSR-VTT}

    \begin{tabular}{l|c c c}
    \toprule
         \textbf{Packet loss} & \textbf{BLEU-4} & \textbf{ROUGE-L} & \textbf{METEOR}  \\
         \textbf{rate (\%)} & & &\\
         \midrule
        \textbf{0.01} & 41.8 & 60.7 & 28.5\\
        \textbf{0.1} & 41.8 & 60.8 &  28.6\\
        \textbf{1} & 40.8& 60.2 & 28.2\\
        \textbf{10} & 35.5 & 56.6 & 25.3 \\
        \textbf{20} & 30.0 & 52.8 & 22.8\\
    \midrule
    \rowcolor{Gray}\textbf{Clean} & 41.5 & 60.9 & 29.0\\
    \bottomrule
    \end{tabular}
    \label{tab:captioning_plr}
\end{table}

\begin{table}
    \centering
    \small
    \caption{UniVL Captioning Performance by Video Length (s), MSR-VTT}
    \begin{tabular}{l|c c c}
    \toprule
         \textbf{Length (s)}  & \textbf{BLEU-4} & \textbf{ROUGE-L} & \textbf{METEOR} \\
         \midrule
        \textbf{0.2} & 19.7 & 48.3 & 19.7 \\
        \textbf{0.5} & 21.3 & 50.8 & 21.2 \\
        \textbf{1} & 23.3 & 53.9 & 23.3\\
        \textbf{2} & 36.5 & 56.4 & 25.1 \\
    \midrule
    \rowcolor{Gray}\textbf{Clean}  & 41.5 & 60.9 & 29.0\\
    \bottomrule
    \end{tabular}
    \label{tab:captioning_length}
\end{table}

\begin{table}
    \centering
    \small
    \caption{UniVL Captioning Performance by Bandwidth, MSR-VTT}

    \begin{tabular}{l|c c c}
    \toprule
         \textbf{Bandwidth} & \textbf{BLEU-4} & \textbf{ROUGE-L} & \textbf{METEOR}\\
         \textbf{(Mbps)} & & &\\
         \midrule
        \textbf{0.12} & 38.5 & 58.2 & 26.3\\
        \textbf{0.5} & 41.1 & 60.2 & 28.1\\
        \textbf{1} & 41.3 & 60.4 &   28.2\\
        \textbf{2} & 41.2 & 60.4 & 28.1\\
        \textbf{3} & 41.2 & 60.4 & 28.1\\
        \textbf{4} & 40.9 & 60.2 & 28.0 \\
        \textbf{5} & 41.1 & 60.2 & 28.0 \\
        \textbf{10} &41.8 & 60.8 &  28.3\\
        \textbf{50} & 41.6 & 60.8 & 28.4\\
        \textbf{100} & 41.6 & 60.7 &  28.4 \\
    \midrule
    \rowcolor{Gray}\textbf{Clean} & 41.5 & 60.9 & 29.0 \\
    \bottomrule
    \end{tabular}
    \label{tab:captioning_bw}
\end{table}

\subsection{Additional Network Parameter Ablations}
\label{appdx:additional_ablations}

We briefly explore the effect of other network conditions on action recognition performance on HMDB51, namely time-varying packet arrivals, delay, and drop-tail queue size. We do not find that these settings non-trivially impact model performance.

\begin{table}
    \centering
    \small
    \caption{3D-ResNet50 Accuracy by Usage of Time-Varying Packet Arrivals, Kinetics-400}
    \begin{tabular}{l | c c}
        \toprule
        \textbf{Link} & \textbf{Time-Var.?} & \textbf{Acc.}\\
        \midrule
            \multirow{ 2}{*}{\textbf{Pantheon-1}} & Y  &  58.1  \\
            &  N  &  58.3  \\
        \midrule
           \multirow{ 2}{*}{\textbf{Pantheon-2}} & Y  &  59.4   \\
             & N   &  59.9  \\     
        \midrule
             \multirow{ 2}{*}{\textbf{Pantheon-3}} & Y  &  59.7   \\
             & N  &  59.3\\
        \midrule
             \rowcolor{Gray}\textbf{Clean} & N/A & 59.9\\
        \bottomrule
    \end{tabular}

   \label{tab:timevar}
\end{table}

\textbf{Time-varying packet arrivals.} Note that time-varying packet arrivals are used in Pantheon-1, -2, -3, the links that yielded the lowest downstream performance after streaming. We study the impact of time-varying links on Pantheon-1, -2, and -3 in isolation, replacing time-varying packet traces used for Pantheon-1, -2, and -3 with time-invariant packet traces of the same bandwidth, before streaming videos from the Kinetics-400 validation split through these links. Ultimately, this has little effect on model performance: our findings are that time-varying packet arrivals do not seem to affect accuracy in excess of 1.8\% (Table~\ref{tab:timevar}). Neither does this choice of network parameter appear to have consistent effects on model performance: time-varying packet arrivals slightly improve accuracy on Pantheon-2 and -3 but slightly hurt accuracy on Pantheon-1. The effect is small and does not explain the accuracy drops seen for Pantheon-1, -2, -3 in the main paper.

\begin{table}
    \centering
    \small
    \caption{3D-ResNet18 Accuracy by Network Delay (ms), HMDB51}
    \begin{tabular}{l | c}
    \toprule
        Delay & Acc.  \\
        \midrule
             \textbf{100} & 55.5\\
             \textbf{50} & 58.8\\
            \textbf{10} & 61.5\\
        \midrule
             \rowcolor{Gray}\textbf{No delay} & 64.8\\
        \bottomrule
    \end{tabular}
   \label{tab:delay}
\end{table}

\textbf{Network Delay.} Table~\ref{tab:delay} shows that increasing network delay appears to have a small but non-negligible effect on downstream model performance. On HMDB51, delays of 100ms resulted in a 9.3\% accuracy drop. At 10ms of delay, however, the accuracy drop was much smaller (3.3\% drop at 10ms delay). While we see higher delay levels on Pantheon-2, -3 (88ms, 130ms), our earlier analysis links the main mechanism of the accuracy drop to low bandwidth, not delay. Ultimately, the correlation between network delay and model performance is unclear, so our analysis is inconclusive.

\begin{table}
    \centering
    \small
    \caption{3D-ResNet18 Accuracy by Queue Size (\# of Packets), HMDB51} 
    \begin{tabular}{l | c}
    \toprule
        Queue Size & Acc.  \\
        \midrule
             \textbf{1} & 4.8\\
             \textbf{5} & 46.4\\
            \textbf{10} & 60.0\\
             \textbf{50} & 62.4\\
             \textbf{100} & 63.0\\
             \textbf{500} &62.9\\
        \midrule
             \rowcolor{Gray}\textbf{Unlimited} & 64.8\\
        \bottomrule
    \end{tabular}

   \label{tab:encoding}
\end{table}

\textbf{Network uplink queue size.} To understand the effect of queue size on model robustness, we evaluate the performance of 3D-ResNet18 on HMDB51 streamed with varying queue sizes. In the real-time video streaming setting, this queue contains network packets that have yet to be decoded from bit-space into pixel-space. We use a droptail queue for these experiments, meaning that arriving packets will be dropped if the queue is already full. Table~\ref{tab:encoding} shows that droptail queue sizes larger than 10 packets have a small effect on model performance, with a 4.8\% decline. However, at queue sizes of 1 and 5 packets, accuracy drops by 16.5\% and 58.1\%, respectively. Note that these two cases represent pathologically small droptail queue sizes, as all Pantheon links have queue sizes of at least 14 (Pantheon-1)

\subsection{Video Length Statistics}
\label{appdx:video_length}

We enumerate the median lengths for all videos streamed via Pantheon. Table~\ref{tab:length_by_bw} shows that as bandwidth limits decrease, video length decreases as well. Since the live-streaming process can change the average frame rate of a video, all lengths are renormalized based on the original frame rate of each video before streaming (length $=$ \# of frames $/$ original frame rate). The number of frames is estimated using FFMpeg.\footnote{Frame counts on near-clean links may appear to slightly exceed original video length. This is because FFMpeg 4.2 sometimes overcounts the number of frames on the corrupted videos by one, which we found by manually counting frames for a small sample of streamed clips. Since SSV2 has an original frame rate of 12, this accounts for the 0.08s (1/12s) increase in video length over the original (Table~\ref{tab:length_by_bw}). Similarly, HMDB51 has an original frame rate of 30, which accounts for the 0.03s (1/30s) increase as well (Table~\ref{tab:length_by_pantheon}).}

Table~\ref{tab:length_by_bw} shows that as bandwidth limits decrease, video length decreases as well for all datasets.
Note that bandwidth limits affect each dataset differently: some datasets suffer from truncation even at 100Mbps, while other datasets suffer no truncation.
However, all datasets experience temporal truncation when the bandwidth limit is sufficiently small.
Table~\ref{tab:length_by_pantheon} shows that most links cause some temporal truncation, but Pantheon-1 through -3 have disproportionate impacts on video length. These findings are consistent with our main study.

\begin{table}
    \centering
    \small
    \caption{Median Video Length (s) by Bandwidth Limit, All Datasets}
    \begin{tabular}{l|c c c c c c c c c c  | c}
    \toprule
    & \multicolumn{10}{c}{\textbf{Bandwidth (Mbps)}}\\
    \textbf{Dataset} & \textbf{0.12} & \textbf{0.5} & \textbf{1} & \textbf{2} & \textbf{3}& \textbf{4} & \textbf{5} & \textbf{10} & \textbf{50} & \textbf{100} & \textbf{Orig.} \\
    \midrule
        \textbf{HMDB51} & 0.07 & 0.43 & 0.60 & 0.77 & 0.93 & 1.17 & 1.50 & 2.50 & 2.63 & 2.63 & 2.67 \\
        \textbf{UCF101} & 1.16 & 4.04 & 4.40 & 4.56 & 4.76 & 4.91 & 5.08 & 5.71 & 6.24 & 6.20 & 6.24\\
        \textbf{Kinetics-Mini} & 1.47 & 3.03 & 5.50 & 6.98 & 7.13 & 7.21 & 7.40 & 8.98 & 10.00 & 10.00 & 10.00 \\
        \textbf{SSV2-Mini} & 0.08 & 0.33 & 0.33 & 0.83 & 1.33 & 1.83 & 2.25 & 3.75 & 4.25 & 4.25 & 4.17\\
        \textbf{OTB-2015} & 1.50 & 6.78 & 9.72 & 11.22 & 11.46 & 11.47 & 11.53 & 11.85 & 12.38 & 12.38 & 13.06\\
        \textbf{DAVIS} & 0.03 & 0.03 & 0.03 & 0.30 & 0.43 & 0.33 & 0.57 & 0.82 & 1.40 & 1.38 & 2.25\\
        \textbf{MSR-VTT} & 3.84 & 10.72 & 10.97 & 11.14 &  11.26 & 11.40 & 11.52 & 11.85 & 12.54 & 13.00 & 13.04 \\
    \bottomrule
    \end{tabular}

    \label{tab:length_by_bw}
\end{table}

\begin{table}
    \centering
    \small
    \caption{Median Video Length (s) by Pantheon Link, All Datasets}
    \begin{tabular}{l|c c c c c c | c}
    \toprule
    & \multicolumn{6}{c}{\textbf{Link}}\\
    \textbf{Dataset} & \textbf{Pan.-1} & \textbf{Pan.-2} & \textbf{Pan.-3} & \textbf{Pan.-4} & \textbf{Pan.-5}& \textbf{Pan.-6} & \textbf{Original} \\
    \midrule
        \textbf{HMDB51} & 0.60 & 0.03 & 1.38 & 2.70 & 2.70  & 2.70 & 2.67 \\
        \textbf{UCF101} & 3.88 & 4.84 & 4.76 & 6.20 & 6.20 & 6.24 & 6.24\\
        \textbf{Kinetics-Mini} & 5.84 & 8.44 & 8.51 & 10.00 & 9.54 & 10.00 & 10.00\\
        \textbf{SSV2-Mini} & 0.08 & 1.58 & 0.42 & 3.33 & 3.33 & 3.33 & 4.17 \\
        \textbf{OTB-2015} & 11.16 & 11.43 & 11.16 & 12.43 & 12.38 & 12.47 & 13.06\\
        \textbf{DAVIS} & 0.60 & 0.05 & 0.17 & 1.40 & 1.48 & 1.40 & 2.25 \\
        \textbf{MSR-VTT} & 10.28 & 11.31 & 11.28 & 13.01 & 13.01 & 13.01  & 13.04 \\
    \bottomrule
    \end{tabular}

    \label{tab:length_by_pantheon}
\end{table}

\section{Additional Defenses for Networking Corruptions}
\label{appdx:defenses}

In addition to data augmentation, we evaluate the following methods as potential defenses against networking corruptions: (1) file checksums, (2) streaming with downsampled video, (3) usage of the HEVC video codec, (4) adversarial training, (5) decreasing training window size, (6) out-of-distribution filtering after streaming, and (7) network corruption-based data augmentation. For defenses involving training a model (adversarial training, decreasing training window size, and data augmentation), if not specified, all training details are the same as those described in Appendix~\ref{appdx:all_details}.
No defense is able to fully restore performance, again reinforcing the need for targeted defenses for the specific artifacts and model failure modes that networking corruptions induce.

\subsection{Methods}

\textbf{Checksums.} We evaluate whether file checksums can be used as a filter for corrupted videos prior to inference. The intuition is that networking corruptions change the underlying bit-representation of each video, suggesting the usage of file checksums to identify corrupted videos. We use the \texttt{md5sum} Linux utility to perform file-integrity checking. 

\textbf{Streaming with downsampled video.} The intuition for downsampling during streaming is that this decreases the bandwidth requirement for streaming videos, potentially mitigating the impact of temporal truncation. Prior to the streaming process, we use the FFMpeg \texttt{scale} filter to downsize the video to 144p, 120p and 96p. We then train a model on clean videos at each resolution, and evaluate on networking-corrupted data with the same resolution. 

\textbf{HEVC codec.} In contrast to the H.264 codec, the HEVC/H.265 codec achieves a better compression ratio, decreasing the bandwidth required for streaming videos. Intuitively, this can potentially mitigate the impact of temporal truncation. Concretely, we simply change the codec used during video transmission to the HEVC/H.265 codec. 

\textbf{Adversarial training.} Adversarial training can potentially mitigate the impact of networking corruptions by limiting worst-case performance within a well-defined set of video perturbations. We augment our training data via the one-step $\ell_\infty$-norm based fast gradient sign method (FGSM) attack~\cite{wong2020fast}. We apply this adversarial perturbation to each video with probability 0.5. 

\textbf{Decreased training window size.} Many video classification training methods, including the training scheme for 3D-ResNet, sample a fixed-size clip from each video for model training. We experiment with decreasing the window size from 16 frames to 8 and 4 frames during model training, with the idea that this can improve robustness to temporal truncation. 

\textbf{Out of distribution filtering.} We use the ODIN out-of-distribution (OOD) detection method~\cite{liang2018enhancing}, which uses a threshold based on transformed softmax score for filtering OOD examples. 
Concretely, this is done using softmax with temperature, which can be computed for each example $\mathbf{x}_i$ as follows:
\begin{equation}
    S_i(\mathbf{x}) = \frac{\exp(f(\*x; \theta) / T)}{\sum_{i=1}^C \exp(f(\*x; \theta) / T)},
\end{equation}
where $T$ is the temperature hyperparameter, $C$ is the number of classes, and $f(\*x; \theta)$ is the model's output. We then select a preset percentile $\alpha$ of the transformed maximum softmax score distribution on the in-distribution data to use as a threshold. During inference, the transformed maximum softmax score is compared to this threshold. Examples with transformed score below this threshold are marked as out-of-distribution and discarded. In our experiments, we choose $T = 1000, \alpha=0.05$. We use HMDB validation split 2 to choose a threshold.

We report accuracy solely over the remaining examples. Furthermore, we also report the AUROC, treating ODIN as a pseudo-classifier for in- vs. out-of-distribution examples. Concretely, AUROC corresponds to the probability that a randomly selected clean example has a higher maximum softmax score than a randomly selected corrupted example.

\textbf{Network Corruption-Based Data Augmentation.} Intuitively, adding noise to the training data resembling networking corruptions may improve model robustness to networking corruptions, as augmentation yields a training distribution closer to the corrupted distribution. We augment the training data by replacing clean videos with their corrupted counterpart with probability 0.5. We train on videos perturbed on one particular link, and evaluate the performance of the resulting model on the validation set corrupted by the same link. Intuitively, this scheme leverages prior knowledge of exact network conditions to simulate networking corruptions that a model may encounter, conditioned on a pre-existing network link. We also attempted to augment training data using networking-corrupted videos from \emph{all} links, but found in practice that this yielded worse performance on both clean data and networking-corrupted videos from individual links. 

To better understand how data augmentation interacts with the artifacts seen under networking corruptions, we also evaluate packet-loss and bandwidth-constraint based augmentation. This allows us to test the utility of data augmentation for visual artifacts and temporal truncation separately.

\subsection{Results}

\subsubsection{File checksums}
\begin{table}
    \centering
    \small
    \caption{Number of Videos with Matching/Non-Matching/Missing Checksums, HMDB51, UCF101, Kinetics-400, and SSV2}
        \vspace{3mm}
    \begin{tabular}{l | c  c c}
    \toprule
        \textbf{Link} & OK & Mismatch & Missing \\
        \midrule
            \textbf{HMDB-Pantheon-1} & 0&	1447&	46\\
            \textbf{HMDB-Pantheon-2} & 0&	1113&	414\\
            \textbf{HMDB-Pantheon-3} &0&	1115&	415\\
            \textbf{HMDB-Pantheon-4} &0&	1520&10\\
            \textbf{HMDB-Pantheon-5} &0&	1510&	20\\
            \textbf{HMDB-Pantheon-6} &0&	1512&	18\\
        \midrule
            \rowcolor{Gray}\textbf{HMDB-Clean}&1530&	0&	0\\
        \midrule
        \midrule
            \textbf{UCF-Pantheon-1} & 0&	3712&	17\\
            \textbf{UCF-Pantheon-2} & 0&	3721&	62\\
            \textbf{UCF-Pantheon-3} &0&	3718&	63\\
            \textbf{UCF-Pantheon-4} &0&	3726&   57 \\
            \textbf{UCF-Pantheon-5} &0&	3715&	68\\
            \textbf{UCF-Pantheon-6} &0&	3749&	34\\
        \midrule
            \rowcolor{Gray}\textbf{UCF-Clean}&3783&	0&	0\\
        \midrule
        \midrule
            \textbf{Kinetics-Pantheon-1} & 0&	15325 &	1972\\
            \textbf{Kinetics-Pantheon-2} & 0&	155787 &	1719\\
            \textbf{Kinetics-Pantheon-3} &0&	15553 &	1744\\
            \textbf{Kinetics-Pantheon-4} &0&	17105 &192\\
            \textbf{Kinetics-Pantheon-5} &0&	17075 &	222\\
            \textbf{Kinetics-Pantheon-6} &0&	17116 &	181\\
        \midrule
            \rowcolor{Gray}\textbf{Kinetics-Clean}&17297&	0&	0\\
        \midrule
        \midrule
            \textbf{SSV2-Pantheon-1} & 0& 8467& 16310\\
            \textbf{SSV2-Pantheon-2} & 0& 7026& 17751\\
            \textbf{SSV2-Pantheon-3} &0&	17870& 6907\\
            \textbf{SSV2-Pantheon-4} &0&	24776& 1\\
           \textbf{SSV2-Pantheon-5} &0&	24772& 5\\
            \textbf{SSV2-Pantheon-6} &0&	24776& 1\\
        \midrule
            \rowcolor{Gray}\textbf{SSV2-Clean}&24777& 0&	0\\
        
        \bottomrule
    \end{tabular}

   \label{tab:ssv2_checksum}
\end{table}
Table~\ref{tab:ssv2_checksum} shows that on all datasets, checksums result in clean videos being discarded indiscriminately, as all streamed videos are flagged as mismatched. This can happen if the streaming process changes the underlying bit-representation of the video without perceptible changes to visual video quality, i.e. under change of video container, different frame rate, or varying encoding parameters (compression quality).

\subsubsection{Streaming with Downsampled Videos}

Downsampling shows promise for mitigating some effects of corruption at the cost of clean-data accuracy. As seen in Table~\ref{tab:downsampled}, evaluating on videos streamed at 144p, clean-data accuracy drops 4.0\%, but accuracy improves on Pantheon-1, -2 and -3 up to 13.7\% (Pantheon-2). At 120p and 96p, accuracy drops further on clean data and Pantheon-1, but improves Pantheon-2 accuracy up to another 3.6\%. All models trained on downsampled videos perform better than the model trained on clean data. However, the model trained to recognize 120p videos performs worse than the model trained for 96p and 144p videos. We retained default hyperparameter settings for all videos; performance may improve with better hyperparameter tuning.

Downsampling may be successful as it lowers the bandwidth requirement for streaming. However, the accuracy drop on clean data suggests that the lower resolution also makes recognizing actions slightly harder. Thus, we conclude that downsampling is promising, but there remains a ~10\% gap to clean-data accuracy. We highlight that it is unclear whether the utilty of downsampling will generalize to other video benchmarks that require finer-grained spatial reasoning; we leave this question to future work.

\begin{table}
    \centering
    \small
    \caption{3D-ResNet18 Accuracy under Networking Corruptions with Downsampled Videos, HMDB51} 
    \begin{tabular}{l|c c c | c}
    \toprule
         \textbf{Link} & 96p & 120p & 144p & Original (240p)\\
    \midrule
         \textbf{Pantheon-1} & 50.9 & 51.4 & 55.2 & 43.3\\
         \textbf{Pantheon-2}  & 57.0 & 48.6 & 50.4 & 37.7\\
         \textbf{Pantheon-3}  & 56.5 & 53.8 & 49.9 & 43.7\\
         \textbf{Pantheon-4}  & 60.6 & 57.6 & 60.4 & 64.2 \\
    \midrule
        \rowcolor{Gray} \textbf{Clean} & 61.4 & 60.2 & 61.5 & 64.8 \\
    \bottomrule
         
    \end{tabular}

    \label{tab:downsampled}
\end{table}

\subsubsection{HEVC Codec}

\begin{table}
    \centering
    \small
    \caption{3D-ResNet18 Accuracy with HEVC/H.265 Codec-based Streaming, HMDB51}
    \begin{tabular}{l|c | c}
    \toprule
         \textbf{Link} & HEVC/H.265 & Original (H.264)\\
    \midrule
         \textbf{Pantheon-1} & 38.8 & 43.3\\
         \textbf{Pantheon-2}  & 37.6 & 37.7 \\
         \textbf{Pantheon-3}  & 28.1 & 43.7\\
         \textbf{Pantheon-4}  & 61.3 & 64.2 \\
    \midrule
        \rowcolor{Gray} \textbf{Clean} & 61.8 & 64.8 \\
    \bottomrule
         
    \end{tabular}

    \label{tab:hevc}
\end{table}

We found that streaming videos using the HEVC codec hurts accuracy on Pantheon 1, -2, and -3 up to 15.6\% (Table~\ref{tab:hevc}). Interestingly, we found that the severity of video truncation worsened when using HEVC, as compared to H.264. For example, the median video length dropped to 0.07s on Pantheon-3 (compare: 1.38s, H.264 on Pantheon-3). We did not tune or experiment with codec parameters beyond the default settings, so we note that accuracy may improve with better codec settings.

\subsubsection{Adversarial Training}

\begin{table}
    \caption{3D-ResNet18 Accuracy with Adversarial Training, HMDB51}
    \centering
    \small
    \begin{tabular}{l|c c c | c}
    \toprule
    \textbf{Link} & $\varepsilon=2$ & $\varepsilon=4$ & $\varepsilon=8$ & Original  \\
    \midrule
        \textbf{Pantheon-1} & 37.6  & 36.1 & 32.6 & 43.3 \\
        \textbf{Pantheon-2} & 29.4 & 31.0 & 25.6 &  37.7 \\
        \textbf{Pantheon-3} & 34.0 & 32.8 & 39.4  & 43.7 \\
        \textbf{Pantheon-4} & 52.2 & 50.6 & 45.3 & 64.2 \\
        \textbf{Pantheon-5} & 52.5 & 50.8 & 45.8 & 64.5 \\
        \textbf{Pantheon-6} & 52.7 & 51.0 & 45.7 & 64.5 \\
    \midrule
        \rowcolor{Gray} Clean & 52.5 & 51.0 & 45.3 & 64.8 \\
    \bottomrule
    \end{tabular}

    \label{tab:adv_results}
\end{table}

Table~\ref{tab:adv_results} shows that adversarial training hurts model accuracy on both clean and networking-corrupted data. This effect is consistent across all Pantheon links. Furthermore, as perturbation budget $\varepsilon$ increases, which corresponds to the maximum variation allowed in an adversarial example, the performance drop worsens as well. Again, this result suggests that the $\ell_p$ norm standard adversarial robustness framework is not applicable to networking corruptions due to its generality. This motivates the exploration of more targeted approaches to mitigating the impact of networking corruptions.

\subsubsection{Decreased Training Window Size}

\begin{table}
    \centering
    \small
    \caption{3D-ResNet18 Accuracy with Varying Training Window Size, HMDB51} 
    \begin{tabular}{l|c c | c}
    \toprule
         \textbf{Link} & 4-frame & 8-frame & Original (16-frame)\\
    \midrule
         \textbf{Pantheon-1} & 19.0 & 24.7 & 43.3\\
         \textbf{Pantheon-2}  & 15.2 & 17.5  & 37.7\\
         \textbf{Pantheon-3}  & 20.0 & 23.7 & 43.7\\
         \textbf{Pantheon-4}  & 23.9 & 31.9 & 64.2\\
    \midrule
        \rowcolor{Gray} \textbf{Clean} & 25.0 & 34.7 & 64.8 \\
    \bottomrule
         
    \end{tabular}

    \label{tab:window_size}
\end{table}

We found that decreasing the video window size used during model training generally harms model performance. However, this lowers clean-data accuracy by 30.3\% when the training video window size is decreased from 16 frames to either 8 or 4 frames (Table~\ref{tab:window_size}). The performance drop suggests that too much temporal information is removed. Clean-data accuracy may improve with further tuning, but we do not observe more robustness to truncation when using smaller training windows, since corrupt accuracy drops into the 20s on Pantheon-1, -2, and -3 in both the 4-frame and 8-frame configurations. Though our initial result is negative, learned frame samplers, which learn to select relevant frames instead of a fixed window, may be more suitable for improving robustness to truncation. We leave this as future work.

\subsubsection{Out of Distribution Filtering}

\begin{table}
    \centering
    \small
    \caption{3D-ResNet18 with OOD Filtering and OOD Detection AUROC, HMDB51}
    
    \begin{tabular}{l|c c | c}
    \toprule
         \textbf{Link} & Acc. Post-OOD & OOD AUROC &  Acc. w/o OOD\\
    \midrule
         \textbf{Pantheon-1} & 52.1 & 81.2 & 43.3 \\
         \textbf{Pantheon-2}  & 51.9 & 82.2 & 37.7 \\
         \textbf{Pantheon-3}  & 60.4 & 75.4 & 43.7\\
         \textbf{Pantheon-4}  & 67.0 & 66.2 & 64.2\\
    \midrule
        \rowcolor{Gray} \textbf{Clean} & 68.8 & 66.1 & 64.8 \\
    \bottomrule
         
    \end{tabular}

    \label{tab:ood}
\end{table}

We find that using the ODIN out-of-distribution detection method as a filter before inference improves accuracy on corrupted HMDB51 by up to 16.7\% (Pantheon-3), leaving a 4\% gap to clean-data accuracy (Table~\ref{tab:ood}). Performance on Pantheon-1 and -2 improves significantly as well, but still lags 12.3\% and 12.5\% respectively behind clean data accuracy. OOD also exhibits moderate discriminatory performance, with AUROC ranging from 75.4 (Pantheon-3) to 82.2 (Pantheon-2). Concretely, when used on the clean dataset, this method discards 11.0\% of clean videos (168 of 1530), which is more than double the expected false positive rate 5.0\%. Thus, ODIN is promising in settings where filtering videos is possible. However, it remains prone to false positives on clean data.

\subsubsection{Networking Corruption-Based Data Augmentation}

 We find that, while data augmentation is unsuccessful in restoring performance, that it can slightly mitigate the impact of severe packet loss, but not temporal truncation. This result suggests that temporal truncation makes classification under networking corruptions inherently more difficult. Intuitively, temporal truncation both decreases the amount of signal available for training, and may also induce label noise if the action of interest is lost. 

\begin{table}
    \centering
    \small
    \caption{3D-ResNet18 Accuracy with Networking-Corruption Augmentation, HMDB51} 
    \begin{tabular}{l|c | c }
    \toprule
    \textbf{Link} & Augmented & No Augmentation \\
    \midrule
        \textbf{Pantheon-1} & 41.9 & 43.3 \\
        \textbf{Pantheon-2}  & 37.2 & 37.7 \\
        \textbf{Pantheon-3} & 42.9 & 43.7 \\
        \textbf{Pantheon-4} & 60.8 & 64.2 \\
    \midrule
      \rowcolor{Gray}\textbf{Clean} & N/A & 64.8\\
    \bottomrule
    \end{tabular}

    \label{tab:aug_results}
\end{table}

\begin{table}
    \centering
    \small
    \caption{3D-ResNet18 Accuracy with Packet Loss Augmentation, HMDB51} 
    \begin{tabular}{l|c | c }
    \toprule
    \textbf{Packet loss} & Augmented & No Augmentation \\
    \textbf{rate (\%)} &  &  \\
    \midrule
        \textbf{0.1} & 59.9 & 62.6  \\
        \textbf{1} & 59.9 & 59.5 \\
        \textbf{10} & 49.6 & 49.2 \\
        \textbf{20} & 36.9 & 33.9 \\
    \midrule
      \rowcolor{Gray}\textbf{Clean} & 59.9  & 64.8 \\
    \bottomrule
    \end{tabular}

    \label{tab:aug_results_plr}
\end{table}

\begin{table}
    \centering
    \small
    \caption{3D-ResNet18 Accuracy with Bandwidth Augmentation, HMDB51}
    
    \begin{tabular}{l|c | c }
    \toprule
    \textbf{Bandwidth} & Augmented & No Augmentation \\
    \textbf{(Mbps)} &  &  \\
    \midrule
        \textbf{0.12} & 35.5 & 39.1 \\
        \textbf{0.5} & 49.6 & 50.1 \\
        \textbf{1} & 51.2 & 54.1 \\
        \textbf{5}  & 59.4 & 60.8\\
        \textbf{10}  & 61.0 & 64.4\\
        \textbf{50}  & 60.8 & 64.5\\
        \textbf{100}  & 60.8 & 64.5\\
    \midrule
      \rowcolor{Gray}\textbf{Clean} & 64.8 & 60.0 \\
    \bottomrule
    \end{tabular}

    \label{tab:aug_results_bw}
\end{table}

\textbf{Data Augmentation.} Table~\ref{tab:aug_results} shows that data augmentation is insufficient to significantly improve corrupt accuracy on HMDB51. At best, on Pantheon-2, data augmentation improves performance by 0.7\%. This suggests that data augmentation is insufficient to improve the robustness of video ML models to networking corruptions. Note that since videos corrupted by Pantheon-5, and -6 were nearly identical to those corrupted by Pantheon-4 on the action recognition task, we computed results for models augmented with networking-corrupted video from Pantheon-1 through -4 only.

\textbf{Packet-loss augmentation.} For packet loss augmentation, we proceed with the data augmentation setup, except that we use videos corrupted at all levels of packet loss (0.1\%, 1\%, 10\%, 20\%). Table~\ref{tab:aug_results_plr} shows that this scheme improves accuracy on highly-corrupted data (3.0\%, 20\% packet loss rate and 0.4\%, 10\% packet loss rate), but improvements taper off as packet loss rate decreases. This suggests that data augmentation might be somewhat effective against visual artifacts. 

\textbf{Bandwidth augmentation.} For bandwidth augmentation, we proceed with the data augmentation setup, except that we use videos corrupted at varying bandwidth limits (0.12, 0.5, 1, 5, 10, 50, 100 Mbps). Table~\ref{tab:aug_results_bw} shows that this scheme consistently harms model performance. Intuitively, bandwidth-based data augmentation results in truncated examples in the training dataset, which may drop important action information. Furthermore, sufficient levels of truncation may actually change the label of the underlying video if the relevant action is dropped from the video, introducing noise into the labels. This can break the assumption in data augmentation that the augmented video retains the same classification label.

\end{document}